\documentclass[11pt]{article}

\usepackage[preprint]{acl}
\usepackage{float}
\usepackage{fancyhdr}

\usepackage{times}
\usepackage{latexsym}
\usepackage{booktabs}
\usepackage{multirow}
\usepackage{array}
\usepackage{colortbl}
\usepackage{xcolor}
\usepackage{amsmath}
\usepackage{makecell}
\usepackage{siunitx}
\usepackage{placeins}
\usepackage{multirow}
\usepackage{graphicx}   
\usepackage{enumitem}
\usepackage{booktabs}
\usepackage{colortbl}   
\usepackage{array}
\usepackage{siunitx}
\usepackage{booktabs,tabularx,array}
\usepackage[T1]{fontenc}

\usepackage[utf8]{inputenc}
\usepackage{adjustbox}
\usepackage{microtype}
\usepackage{array}
\newcolumntype{L}[1]{>{\raggedright\arraybackslash}p{#1}}
\usepackage{siunitx}
\usepackage[table]{xcolor}
\usepackage{pifont}
\newcommand{\cmark}{\ding{51}}
\newcommand{\xmark}{\ding{55}}
\definecolor{hl}{RGB}{255,245,220}

\usepackage{graphicx}

\title{EviRank: Structured Relevance Evidence for Multimodal Image Re-ranking}

\author{
  Enjun Du\textsuperscript{1,2,3},
  Siyi Liu\textsuperscript{1},
  Zirong Chen\textsuperscript{1,2},
  Xinyu Zuo\textsuperscript{2},
  Jinwen Luo\textsuperscript{2},
  Ruiwen Tao\textsuperscript{2},
  Lisheng Duan\textsuperscript{2},
  Haijin Liang\textsuperscript{2},
  Jin Ma\textsuperscript{2},
  Junfu Pu\textsuperscript{2},
  Yongqi Zhang\textsuperscript{2}
}

\begin{document}
\pagestyle{fancy}
\fancyhf{}
\lhead{\small Yuanbao Technical Report}
\rhead{\small Tencent}
\cfoot{\thepage}
\setlength{\headheight}{14pt}
\thispagestyle{fancy}

\twocolumn[{
\begin{minipage}{\textwidth}
  \noindent
  \includegraphics[width=0.18\linewidth]{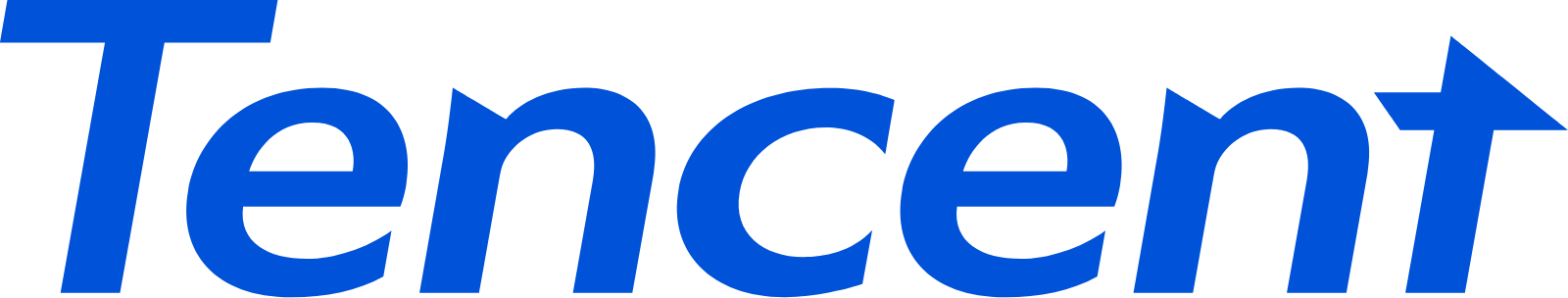}
  \hfill
  \includegraphics[height=1.25cm]{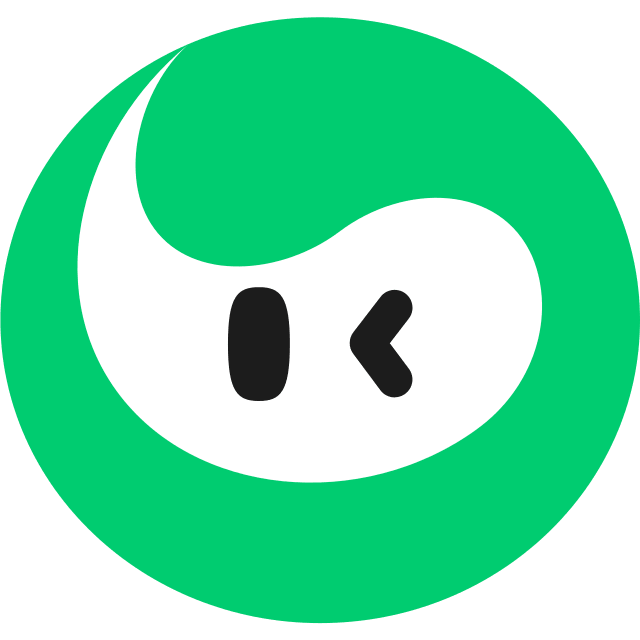}

  \vspace{0.45cm}
  \begin{center}
  {\Huge\bfseries
  EviRank: Structured Relevance Evidence\\[0.2em]
  for Multimodal Image Re-ranking\par}

  \vspace{0.5cm}
  {\large
  Enjun Du\textsuperscript{1,2,3,\textdagger} \quad
  Siyi Liu\textsuperscript{1} \quad
  Zirong Chen\textsuperscript{1,2,\textdagger} \quad
  Xinyu Zuo\textsuperscript{2} \quad
  Jinwen Luo\textsuperscript{2} \quad
  Ruiwen Tao\textsuperscript{2}\\[0.25em]
  Lisheng Duan\textsuperscript{2} \quad
  Haijin Liang\textsuperscript{2} \quad
  Jin Ma\textsuperscript{2} \quad
  Junfu Pu\textsuperscript{2} \quad
  Yongqi Zhang\textsuperscript{1,*}\par}

  \vspace{0.3cm}
  {\small
  \textsuperscript{1}The Hong Kong University of Science and Technology (Guangzhou)\\[0.1em]
  \textsuperscript{2}Tencent Yuanbao \qquad
  \textsuperscript{3}The University of Hong Kong\\[0.18em]

  \texttt{\{enjundu, yongqizhang\}@hkust-gz.edu.cn}\par}
  \end{center}
  \vspace{0.25cm}
\end{minipage}
}]
\thispagestyle{fancy}

\begingroup
\renewcommand{\thefootnote}{\textdagger}
\footnotetext{Work done during an internship at Tencent.}
\renewcommand{\thefootnote}{*}
\footnotetext{Corresponding author.}
\endgroup




\begin{abstract}
Real-world image search queries are multimodal and compositional:
``find this shirt in pink'' specifies an entity to retain, an
attribute to modify, and context to ignore. Yet existing
re-rankers either compress such multi-faceted relevance into an
opaque embedding or rely on free-form chain-of-thought that
easily omits or hallucinates fine-grained constraints. Drawing on
rubric- and checklist-based evaluation from NLP, we recast
multimodal image re-ranking as a \textbf{semantic constraint
satisfaction} problem and propose \textbf{EviRank}, which parses
any query---text-only, image-only, or composed---into a unified
\emph{evidence package}: typed criteria across six semantic
slots (e.g., entities, attributes, relations), each labelled
\emph{required}, \emph{forbidden}, or \emph{ignorable}.
Re-ranking then reduces to \textbf{evidence-conditioned
verification}, combining deterministic rubric scoring and
evidence-grounded listwise comparison in a single training-free
procedure. The explicit evidence can further serve as structured
supervision for optionally distilling a lightweight student.
Across five benchmarks spanning text-to-image, image-to-image,
and composed image retrieval, EviRank achieves state-of-the-art
performance, and the distilled student preserves over 90\% of
the teacher's capability at substantially lower cost. Code is available at \href{https://github.com/EnjunDu/EviRank}{https://github.com/EnjunDu/EviRank}.
\end{abstract}



\section{Introduction}\label{sec:intro}

Real-world image search queries are rarely single-faceted. A user
asking ``find \emph{this shirt} in \emph{pink}'' is implicitly
specifying multiple semantic constraints at once: the system must
\emph{retain} the entity ``shirt'' together with its visual
identity (cut, pattern, fit), \emph{modify} a specific attribute
(color $\rightarrow$ pink), and \emph{ignore} incidental factors
such as background or lighting in the reference
image~\cite{Xing2025ConTextCIR,Tang2025ReasonBeforeRetrieve,Tian2025CCIN}.
Such queries naturally combine information from multiple
modalities---text, image, or both---and constrain the target along
several semantic dimensions, such as \emph{entities},
\emph{attributes}, \emph{actions}, \emph{relations}, and
discriminative \emph{details}. Whether the query is purely
textual, purely visual, or a mixture of the two, the underlying
problem is the same: deciding whether each candidate image
satisfies the set of fine-grained semantic constraints that the
query imposes.

Existing multimodal re-rankers, however, do not explicitly model
these constraints. \emph{Representation-centric}
approaches---ranging from classical dual-encoder
embeddings~\cite{frome2013devise} to recent MLLM-based
scorers~\cite{Liu2025LamRA,Huynh2025CoLLM,Sun2025CIRLVLM}---collapse
all semantic dimensions of a query into a single dense vector.
Such an opaque representation cannot tell which evidence should
be preserved, modified, or ignored, and offers no handle on
fine-grained constraint violations.
\emph{Reasoning-centric} approaches~\cite{Sun2025CoTMR,Wu2025CoTRR}
instead ask an MLLM to reason about relevance in free-form
chain-of-thought (CoT). While more expressive, this free-form
reasoning remains \emph{unstructured} natural language, and
exhibits three concrete failure modes.
First, it \emph{omits} implicit constraints: when a user says
``find this shirt in pink,'' a CoT trace typically discusses the
color change while silently dropping the ``preserve the original
cut and pattern'' constraint.
Second, it \emph{hallucinates} evidence, attributing visual
properties the model has not actually verified.
Third, its semantic coverage is \emph{inconsistent across
queries}, so completeness and verifiability of the relevance
judgement cannot be guaranteed without a formal structure
specifying which dimensions must be examined.
These failure modes share a common root: multimodal image
re-ranking is fundamentally a \textbf{semantic constraint
satisfaction} problem, not a generic similarity-matching
problem, and methods that do not represent constraints
explicitly cannot systematically verify them.

We therefore propose to represent multimodal relevance as a
\textbf{structured Evidence Frame}. Instead of free-form
reasoning, we first \emph{parse the query into typed evidence
statements}---\textsc{required}, \textsc{forbidden}, and
\textsc{ignorable}---organized along six semantic slots
(entities, attributes, actions, relations, scene, and key
details; full definitions in Section~\ref{sec:problem}). The
three labels mirror well-studied semantic relations
(entailment, contradiction, and invariance, respectively;
Section~\ref{sec:problem}). Since the Evidence Frame is
modality-agnostic, text-only, image-only and composed queries
share a single evidence interface. Re-ranking then reduces to
\emph{verifying} which constraints each candidate satisfies or
violates---a slot-wise, language-grounded judgement that is
transparent and deterministically aggregable.

Building on this representation, \textbf{EviRank} re-ranks
top-$K$ candidates in two stages---\textbf{evidence mining}
and \textbf{evidence verification}---both training-free. The
same slot-wise satisfaction scores and hard-pair distinctions
also serve as naturally decomposable supervision for distilling
a lightweight student re-ranker. Our core contributions are:

\begin{itemize}[leftmargin=*,nosep]
    \item \textbf{A new formulation.} We formulate multimodal
    image re-ranking as \emph{evidence-conditioned semantic
    verification}, where text-only, image-only and composed
    queries are converted into a shared
    \emph{Evidence Frame} of typed relevance criteria
    (required / forbidden / ignorable, instantiated as
    entailment / contradiction / invariance).
    \item \textbf{Evidence mining and evidence verification.}
    We design a training-free, evidence-conditioned re-ranking
    procedure that mines slot-wise required, forbidden, and
    ignorable evidence, and verifies it through a deterministic
    rubric grounded in the same evidence.
    \item \textbf{Evidence as supervision.} We show that the
    structured evidence and its slot-wise verification signals
    form a naturally decomposable supervision source, which can
    be distilled into a compact student re-ranker.
    \item \textbf{Experiments.}
    State-of-the-art results across five benchmarks, with
    analysis showing the structured evidence is robust,
    interpretable, and contributes beyond free-form augmentation.
\end{itemize}

\section{Related Work}
\label{sec:related}

\paragraph{Implicit Relevance Representations for Multimodal Retrieval.}
A first family encodes queries and candidate images into a joint
embedding space and ranks by
similarity~\cite{frome2013devise,Liu2024DualEncoderReRank,Levy2024CASE,Bai2024SPRC};
recent MLLM-based re-rankers continue this paradigm by distilling
multimodal capability into improved embeddings or learned scoring
heads~\cite{Liu2025LamRA,Huynh2025CoLLM,Sun2025CIRLVLM,Lin2025MMEmbed,Chen2024RAGVL},
while a complementary line reformulates retrieval as
document-identifier generation~\cite{tay2022dsi,wang2022nci,li2024genir}.
All of them represent relevance \emph{implicitly}---a scalar
similarity, an opaque embedding, or a learned ID
distribution---and cannot tell which semantic dimensions of a
query a candidate satisfies or violates. Our work is orthogonal:
we operate on top of any such retriever and replace its implicit
score with explicit, slot-wise evidence verification.

\paragraph{Free-form Reasoning for Image Re-ranking.}
A second family augments MLLM-based re-ranking with CoT
reasoning~\cite{Sun2025CoTMR,Wu2025CoTRR,Tang2025OSrCIR,Luo2025ImageScope}.
More expressive than a single similarity score, the reasoning
trace nonetheless remains \emph{unstructured natural language} and
exhibits the omission, hallucination, and inconsistent-coverage
failures discussed in Section~\ref{sec:intro}. EviRank instead
asks the teacher for a typed Evidence Frame (required / forbidden
/ ignorable over six semantic slots) that can be verified,
aggregated, and distilled.

\begin{table}[t]
\centering
\caption{Novelty delineation against the closest reasoning-based
re-rankers. \cmark: reported as an explicit, fixed and reusable
representational mechanism; $\circ$: partially or implicitly
present; \xmark: not reported as such a mechanism---which does not
deny that the underlying MLLM may implicitly possess the
capability.}
\label{tab:novelty}
\resizebox{\columnwidth}{!}{%
\footnotesize
\setlength{\tabcolsep}{3pt}
\renewcommand{\arraystretch}{0.95}
\begin{tabular}{@{}L{3.0cm}cccc@{}}
\toprule
\textbf{Explicit mechanism}
& \makecell{\textbf{Image}\\\textbf{Scope}}
& \makecell{\textbf{CoT}\\\textbf{RR}}
& \makecell{\textbf{CoT}\\\textbf{MR}}
& \textbf{Ours} \\
\midrule
Fixed, reusable slot taxonomy      & $\circ$ & \xmark & \xmark & \cmark \\
Typed \textsc{forbidden} evidence  & \xmark  & \xmark & $\circ$ & \cmark \\
Typed \textsc{ignorable} evidence  & \xmark  & \xmark & \xmark & \cmark \\
One frame for T$\to$I/I$\to$I/CIR  & $\circ$ & \xmark & \xmark & \cmark \\
Deterministic auditable scoring    & \xmark  & \xmark & \xmark & \cmark \\
Teacher-free distilled student     & \xmark  & \xmark & \xmark & \cmark \\
\bottomrule
\end{tabular}}
\end{table}

\paragraph{Novelty delineation.}
Our claim is \emph{representational} rather than about what MLLMs
can reason about. ImageScope decomposes a query into a
multi-granularity hierarchy, CoTRR into five reasoning components
and CoTMR into predefined verification subtasks, yet in all three
the intermediate product is free-form text whose fields are
neither typed nor guaranteed to recur across queries. EviRank
fixes a six-slot taxonomy and attaches an explicit polarity to
every statement, which is what makes deterministic aggregation
(Eqs.~\ref{eq:slot_rates}--\ref{eq:rubric}), per-slot auditing and
reuse as typed supervision possible
(Table~\ref{tab:novelty}; formulation-level comparison with
ImageScope in Appendix~\ref{appendix:imagescope_compare}).

\paragraph{Structured Criteria for Semantic Verification.}
The NLP community has increasingly turned to structured rubric-
and checklist-based evaluation, where complex semantic judgements
are decomposed into independently checkable
criteria~\cite{Liu2023GEval,Cook2024TICK,Lee2024CheckEval,Wei2025RocketEval},
and rubric criteria are further used as RL reward
signals~\cite{Zhou2025RuscaRL,2025HealthBench}. A closely related
line decomposes complex \emph{claims} into verifiable sub-units
for fact checking and faithfulness
evaluation~\cite{Thorne2018FEVER,Min2023FActScore,Golovneva2023ROSCOE},
while a further line makes the intermediate structure itself
explicit and inspectable---as a provenance-constrained evidence
ledger, as reasoning paths or relational digraphs over a
graph~\cite{du2026ledgermind,du2025mokgr,zhang2026neural,zhang2023adaprop},
or as a structure shared between a neural reasoner and an
LLM~\cite{liu2025dual}.
EviRank imports the same \emph{decompose-then-verify} principle
into multimodal image re-ranking: a multimodal query becomes typed
evidence statements over six semantic slots, and relevance is
established by slot-wise, language-grounded verification rather
than by opaque similarity or free-form CoT.

\begin{figure*}[htp]
  \centering
  \includegraphics[width=1.98\columnwidth]{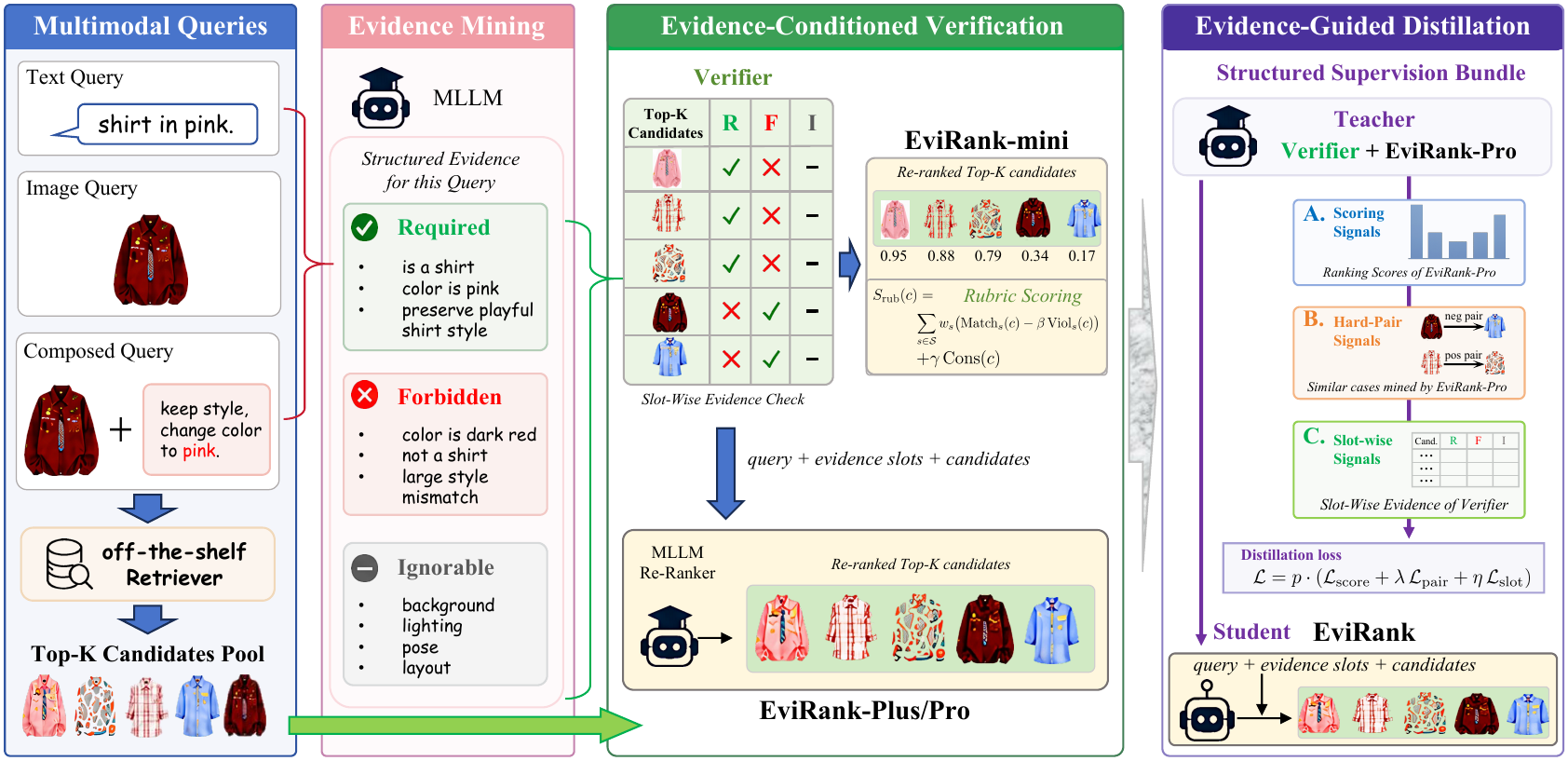}
  \caption{\textbf{Overview of \textbf{EviRank}.}
  A multimodal query (text-only, image-only, or composed) and
  top-$K$ candidates from any retriever are first parsed by an
  MLLM teacher into a structured Evidence Frame of typed
  \emph{required} / \emph{forbidden} / \emph{ignorable}
  statements over six semantic slots. Candidates are then
  re-ranked by evidence-conditioned verification (deterministic
  rubric + MLLM listwise refinement). The same evidence yields a
  structured supervision bundle that is distilled into a
  lightweight student; at inference the student consumes only
  the raw query and candidate image.}
  \label{fig:main_figure}
\end{figure*}

\section{Method}
\label{sec:method}

We present \textbf{EviRank}
(Figure~\ref{fig:main_figure}), which recasts multimodal image
re-ranking as \emph{semantic constraint satisfaction}: a query
is parsed into a unified package of structured evidence
criteria, and candidates are re-ranked by checking which
criteria they satisfy or violate.
Section~\ref{sec:problem} defines the setting and evidence
representation; Sections~\ref{sec:mining}
and~\ref{sec:verify} describe evidence mining and
evidence-conditioned verification and re-ranking; and
Section~\ref{sec:distill} shows how the same evidence serves as
decomposable supervision for distilling a lightweight
re-ranker.

\subsection{Setting and Evidence Representation}
\label{sec:problem}

\paragraph{Setting.}
We consider multimodal image re-ranking. A query $q$ may consist
of text, an image, or both---without loss of generality, we write
$q=(t,I_{\text{ref}})$ where either component may be empty. Given
$q$ and a top-$K$ candidate set
$\mathcal{C}_K=\{c_1,\dots,c_K\}$ produced by any off-the-shelf
retriever, the goal of re-ranking is to output a permutation
$\pi$ over $\mathcal{C}_K$ that better reflects semantic relevance.

\paragraph{From queries to evidence.}
Our central observation is that, regardless of how $q$ is
expressed, it implicitly defines a set of \emph{semantic
constraints} on the target image: some elements must be present,
others must not, and many are simply irrelevant. We therefore
represent relevance not by a similarity score, but by a typed
\textbf{evidence package}
\begin{equation}
\mathcal{E}(q) = \bigl(g,\;\{(R_s, F_s, I_s)\}_{s\in\mathcal{S}}\bigr),
\label{eq:evidence}
\end{equation}
where $g$ is a concise global summary of the query intent and
$\mathcal{S}$ is a lightweight default schema of six semantic
slots,

\[
\mathcal{S} =
\left\{
\begin{aligned}
&\textsc{Entities}, \textsc{Attributes}, \textsc{Actions},\\
&\textsc{Relations}, \textsc{Scene}, \textsc{KeyDetails}
\end{aligned}
\right\}.
\]

Slots are allowed to remain empty when the query does not
express the corresponding factor (e.g., a still-life caption
typically populates \textsc{Entities} and \textsc{Attributes}
but leaves \textsc{Actions} and \textsc{Relations} empty), and
the schema can be extended with additional slots for
specialized domains without changing the verification
procedure. Each slot $s$ contributes three lists of short, checkable
sentences: \emph{required} constraints $R_s$ that must hold,
\emph{forbidden} constraints $F_s$ that must not hold, and
\emph{ignorable} constraints $I_s$ that should not influence the
decision (Appendix~\ref{appendix:sql} draws an analogy between
this typing and classical SQL constraints).

\paragraph{Unifying heterogeneous query forms.}
A key property of this representation is that it is
\emph{modality-agnostic}: although users may issue queries in very
different forms, the resulting evidence package always lives in
the same schema. A purely textual query becomes evidence whose
required slots reflect what the user verbalizes and whose
ignorable slots absorb under-specified context. A purely visual
query is first captioned and then converted to evidence in which
the reference image's salient content becomes the required
constraints. A composed query (image $+$ modification text) is
mapped to evidence by treating the reference image as a source of
\emph{preserved} required constraints (e.g., shirt cut and
pattern) and the modification text as a source of \emph{updated}
required constraints together with newly induced forbidden
constraints (e.g., original color $\to$ pink, so ``the shirt is
of the original color'' becomes forbidden). After this normalization, all subsequent stages of EviRank
operate on the same evidence schema, so heterogeneous query
forms can be handled through a shared evidence-verification
interface rather than paradigm-specific re-ranking pipelines.

\subsection{Query-to-Evidence Mining}
\label{sec:mining}

Building the evidence package $\mathcal{E}(q)$ proceeds in two
training-free steps performed by a single multimodal model
(referred to as the MLLM teacher throughout this paper).

\paragraph{Query normalization.}
Different query forms expose different amounts of information.
Text queries are explicit but often under-specified; image queries
are visually rich but lexically unanchored; composed queries
entangle both. We therefore first normalize $q$ into a textual
representation that exposes its semantic content in a uniform
way. For text queries, we \emph{verbalize} $t$ into $\tilde{t}$
that makes implicit visual details (identity, scene, salient
attributes) explicit, while preserving the original intent.
For image queries, we generate a concise caption together with key
objects. For composed queries, we combine the reference caption
with the modification text and explicitly mark which aspects
should be \emph{preserved} versus \emph{modified}. This step is
training-free; it serves purely as semantic enrichment for
downstream evidence extraction, \emph{not} as an additional
retrieval pass.

\paragraph{Evidence extraction.}
Given the normalized query, the MLLM emits the full evidence
package $\mathcal{E}(q)$ in a single structured pass. For each
slot $s\in\mathcal{S}$, the model produces 2--4 short affirmative
sentences for each of $R_s$, $F_s$, and $I_s$. Critically, even
forbidden constraints are written as positive statements that
describe what an \emph{incorrect} match would look like (e.g.,
``the shirt is red''), so that every constraint can be compared
to a candidate by the same matching operator and yields
sign-consistent scores during verification. Ignorable constraints
explicitly enumerate non-discriminative variations (e.g., minor
shade shifts, background lighting) that should be tolerated
rather than penalized. The full prompts used for query
normalization and evidence extraction are in
Appendix~\ref{appendix:prompts}. The six slots cover the principal semantic axes of multimodal
queries (\emph{entities}, \emph{attributes}, \emph{actions},
\emph{relations}, \emph{scene}, \emph{key details}) and are
empirically near-orthogonal (mean pairwise SBERT similarity
0.18; full slot-semantics analysis in
Appendix~\ref{appendix:slot_analysis}); the schema is a default
rather than a hard requirement (cf.\ Section~\ref{sec:problem}).

\subsection{Evidence-Conditioned Verification and Re-ranking}
\label{sec:verify}

Once $\mathcal{E}(q)$ is mined, re-ranking reduces to
\emph{verifying} which constraints each candidate satisfies. We
combine a deterministic rubric for stable, interpretable per-
candidate scoring with an evidence-grounded listwise step for
global, comparative judgement; both operate on the same evidence.

\paragraph{Deterministic rubric scoring.}
Let $m(\cdot,c)\in[0,1]$ be a matching operator that measures
whether a constraint sentence holds for $c$. We instantiate $m$
either as embedding-based cosine similarity using a pretrained
vision--language encoder, or as a teacher-graded score. The slot
satisfaction and violation rates are
\begin{equation}
\begin{aligned}
\text{Match}_s(c)
&= \frac{1}{|R_s|}\sum_{r\in R_s} m(r,c), \\
\text{Viol}_s(c)
&= \frac{1}{|F_s|}\sum_{f\in F_s} m(f,c).
\end{aligned}
\label{eq:slot_rates}
\end{equation}
and the rubric score aggregates them with a forbidden penalty and
an optional cross-modal consistency term:
\begin{equation}
\begin{aligned}
S_{\text{rub}}(c)
&= \sum_{s\in\mathcal{S}} w_s
\bigl(\text{Match}_s(c)-\beta\,\text{Viol}_s(c)\bigr) \\
&\quad + \gamma\,\text{Cons}(c).
\end{aligned}
\label{eq:rubric}
\end{equation}
Slot weights $w_s$ are uniform; $\beta=0.75$ controls the strength
of forbidden penalties; and $\text{Cons}(c)$, used only when the
query has both text and image components, measures alignment
between the expanded query text $\tilde t$ and $c$ via a
pretrained encoder. Ignorable constraints $I_s$ are not scored
directly; instead they mask required constraints whose embedding
is too close to any ignore statement, preventing background or
stylistic cues from dominating the score.

\paragraph{Evidence-grounded listwise refinement.}
The rubric is stable and interpretable but evaluates candidates
independently and therefore cannot easily resolve close calls
between visually similar candidates. We address this with an evidence-grounded listwise step. The
teacher is shown the query, the global intent $g$ together with
the \emph{required} and \emph{forbidden} constraints of
$\mathcal{E}(q)$, and the candidates jointly, and is asked to
return a permutation in which required constraints raise rank
when clearly satisfied and forbidden constraints heavily
penalize salient mismatches. \emph{Ignorable} constraints are
not exposed to the listwise prompt; they act as an invariance
mask inside the deterministic rubric only, keeping the prompt
compact and avoiding asking the teacher to reason over
non-discriminative details. In practice, we sort candidates by
$S_{\text{rub}}$ and invoke the teacher only to verify and locally
re-order the top-$M$ ($M{=}5$ by default), which substantially
reduces cost while preserving accuracy. For larger candidate
pools, candidates are processed in batches and merged in a
divide-and-conquer fashion (details in Appendix~\ref{app:listwise_batch}).

\paragraph{A by-product: structured supervision.}
Beyond the final ranking, the listwise step naturally produces a
JSON-formatted artifact containing per-candidate relevance scores,
a self-assessed confidence, and a short list of \emph{hard pairs}
(closest competitors with brief distinguishing reasons). These
auxiliary signals are not needed for inference but are reused as
supervision in the optional distillation stage
(Section~\ref{sec:distill}).

\subsection{Evidence as Supervision: Structured Distillation}
\label{sec:distill}

Beyond verification, the structured evidence package
$\mathcal{E}(q)$ also serves as a \emph{structured supervision
signal}. Unlike free-form CoT traces, which a smaller student
must imitate verbatim, every component of our pipeline is typed
and decomposable, letting us turn the teacher's behaviour into
a precisely-targeted supervision bundle for the student.

For each query, the verification stage yields a query-conditioned
\emph{supervision bundle} comprising
(i) the evidence package $\mathcal{E}(q)$,
(ii) per-candidate slot-wise satisfaction and violation rates,
(iii) calibrated listwise scores $\mathbf{r}=(r_1,\dots,r_K)$,
(iv) a self-assessed confidence $p\in[0,1]$, and
(v) a set of hard pairs $\mathcal{H}=\{(i,j)\}$ with
distinguishing constraints. Every component has a fixed schema and a clear semantic role,
making the supervision transferable to a smaller model.
\paragraph{Student input: evidence as training-time supervision only.}
We distil this evidence-grounded behaviour into a lightweight
student re-ranker. Crucially, at \emph{inference time} the
student takes only the raw query (text and/or reference image)
and a candidate image as input, and predicts a relevance score
$s_i \in [0,1]$. The evidence package $\mathcal{E}(q)$,
slot-wise satisfaction labels, listwise scores, hard pairs, and
confidence are used \emph{only as supervision during training}
and are never required at deployment. The student therefore
runs without any teacher MLLM, evidence generator, or
runtime evidence cache.

\paragraph{Training objectives.}
Three objectives mirror the three structured components of the
bundle. Let $\mathbf{s}=(s_1,\dots,s_K)$ be the student scores
for the same list of $K$ candidates. \emph{Score distillation}
standardizes the teacher scores per query,
$\hat{r}_i=(r_i-\mu_r)/(\sigma_r+\epsilon)$ with $\mu_r,\sigma_r$
the mean and standard deviation of $\mathbf{r}$ and
$\epsilon=10^{-8}$, and matches the two listwise distributions at
temperature $\tau$:
\begin{equation*}
\mathcal{L}_{\text{score}}=\mathrm{KL}\!\bigl(\mathrm{softmax}(\hat{\mathbf{r}}/\tau)\,\|\,\mathrm{softmax}(\mathbf{s}/\tau)\bigr).
\end{equation*}
\emph{Hard-pair supervision} imposes a margin $\delta$ on every
teacher-flagged pair $(i,j)\in\mathcal{H}$ in which $i$ should
outrank $j$, weighted inversely to the teacher's score gap so
that closely-scored competitors dominate the gradient:
\begin{align*}
w_{ij}&=\tfrac{1}{|r_i-r_j|+\epsilon},\\
\mathcal{L}_{\text{pair}}&=\!\!\sum_{(i,j)\in\mathcal{H}}\!\! w_{ij}
\max\bigl(0,\;\delta-(s_i-s_j)\bigr).
\end{align*}
\emph{Slot-wise supervision} is optional: when slot-wise labels
$y_{s,i}=\mathbf{1}[\text{Match}_s(c_i)>\theta]$ are available,
a lightweight auxiliary head predicts $\hat{y}_{s,i}$ for each
slot $s\in\mathcal{S}$ and is trained by binary cross-entropy,
\begin{equation*}
\resizebox{\columnwidth}{!}{$
\mathcal{L}_{\text{slot}}=-\tfrac{1}{K|\mathcal{S}|}
\!\!\sum_{i,s}\!\!\bigl[y_{s,i}\log\hat{y}_{s,i}+(1{-}y_{s,i})\log(1{-}\hat{y}_{s,i})\bigr],
$}
\end{equation*}
so that the student internalizes \emph{which} constraints make a
candidate a good match rather than only its position. The three
terms are combined and modulated by the teacher's confidence $p$,
so that uncertain supervision contributes less:
\begin{equation}
\mathcal{L}=p\cdot\bigl(\mathcal{L}_{\text{score}}
+\lambda\,\mathcal{L}_{\text{pair}}
+\eta\,\mathcal{L}_{\text{slot}}\bigr).
\label{eq:total_loss}
\end{equation}
The auxiliary head is discarded after training, so inference
cost is unchanged. Values of $\tau,\delta,\lambda,\eta$ and
further implementation details are given in
Appendix~\ref{appendix:distill_loss}.

\section{Experiments}

\begin{table}[t]
\centering
\caption{Retrieval performance (\%) with different backbones on Flickr30k.}
\label{tab:flickr30k}
\resizebox{\columnwidth}{!}{%
\fontsize{7.0}{8.0}\selectfont
\setlength{\tabcolsep}{3pt}
\renewcommand{\arraystretch}{0.90}
\begin{tabular}{
  L{0.75cm}
  L{1.4cm}
  S[table-format=2.1]
  S[table-format=2.1]
  S[table-format=2.1]
  S[table-format=2.1]
}
\toprule
\textbf{VLM} & \textbf{Methods}
& {\textbf{R@1}} & {\textbf{R@5}} & {\textbf{R@10}} & {\textbf{MRR@5}} \\
\midrule
& No ReRank    & 84.0 & 95.3 & 96.9 & 89.7 \\
& ImageScope   & 84.8 & 95.6 & 97.0 & 90.1 \\
& CoTRR        & 85.6 & 96.2 & 97.3 & 90.3 \\
& CoTMR        & 85.9 & 96.8 & 97.6 & 90.9 \\
\rowcolor{hl} & EviRank-mini & 84.9 & 96.5 & 97.5 & 91.0 \\
\rowcolor{hl} & EviRank      & 86.7 & 96.9 & 98.0 & 91.1 \\
\rowcolor{hl} & EviRank-plus & 87.2 & 97.8 & 98.4 & 92.2 \\
\rowcolor{hl}\multirow{-8}{*}{\rotatebox[origin=c]{90}{EVA-CLIP-18B}}
& EviRank-pro  & 88.0 & 97.7 & 98.7 & 92.5 \\
\midrule
& No ReRank    & 67.1 & 89.0 & 90.0 & 75.8 \\
& AFS          & 72.4 & 91.3 & 91.8 & 80.1 \\
& Explicit     & 72.5 & 93.7 & 93.9 & 80.9 \\
& ImageScope   & 78.8 & 92.6 & 95.6 & 80.2 \\
& CoTRR        & 84.6 & 94.7 & 96.0 & 80.9 \\
& CoTMR        & 84.7 & 94.8 & 96.3 & 81.1 \\
\rowcolor{hl} & EviRank-mini & 85.2 & 94.0 & 96.4 & 82.7 \\
\rowcolor{hl} & EviRank      & 86.2 & 95.0 & 96.7 & 86.8 \\
\rowcolor{hl} & EviRank-plus & 87.1 & 96.3 & 97.1 & 89.9 \\
\rowcolor{hl}\multirow{-10}{*}{\rotatebox[origin=c]{90}{CLIP-ViT-B/32}}
& EviRank-pro  & 87.2 & 97.2 & 97.9 & 90.2 \\
\midrule
& No ReRank    & 72.7 & 91.2 & 92.4 & 80.0 \\
& AFS          & 78.4 & 94.3 & 95.2 & 84.6 \\
& Explicit     & 74.6 & 93.3 & 94.2 & 82.0 \\
& ImageScope   & 78.1 & 91.0 & 93.3 & 82.2 \\
& CoTRR        & 83.8 & 93.2 & 94.2 & 82.8 \\
& CoTMR        & 84.5 & 93.9 & 94.3 & 83.0 \\
\rowcolor{hl} & EviRank-mini & 83.6 & 92.4 & 93.4 & 82.3 \\
\rowcolor{hl} & EviRank      & 85.0 & 94.3 & 95.4 & 84.9 \\
\rowcolor{hl} & EviRank-plus & 86.7 & 94.6 & 96.1 & 85.4 \\
\rowcolor{hl}\multirow{-10}{*}{\rotatebox[origin=c]{90}{CLIP-ViT-L/14}}
& EviRank-pro  & 88.7 & 95.5 & 96.9 & 87.1 \\
\midrule
& No ReRank    & 86.8 & 97.5 & 98.0 & 91.5 \\
& AFS          & 89.0 & 98.2 & 98.3 & 93.1 \\
& Explicit     & 88.6 & 98.4 & 98.5 & 92.8 \\
& ImageScope   & 88.8 & 98.2 & 98.3 & 92.8 \\
& CoTRR        & 88.9 & 98.3 & 98.4 & 93.0 \\
& CoTMR        & 89.2 & 98.4 & 98.9 & 93.4 \\
\rowcolor{hl} & EviRank-mini & 89.3 & 98.6 & 98.9 & 93.3 \\
\rowcolor{hl} & EviRank      & 91.3 & 98.6 & 99.0 & 93.6 \\
\rowcolor{hl} & EviRank-plus & 93.5 & 99.0 & 99.2 & 94.2 \\
\rowcolor{hl}\multirow{-10}{*}{\rotatebox[origin=c]{90}{BLIP-2}}
& EviRank-pro  & 95.6 & 99.2 & 99.4 & 95.6 \\
\bottomrule
\end{tabular}
}
\end{table}

\subsection{Experimental Setup}
\label{sec:setup}

\paragraph{Datasets.}
We evaluate \textbf{EviRank} across five benchmarks spanning three retrieval
paradigms: \textbf{T$\rightarrow$I} (MS~COCO~\cite{Lin2014COCO},
Flickr30k~\cite{Plummer2015Flickr30k}), \textbf{I$\rightarrow$I} (Stanford Online
Products (SoP)~\cite{OhSong2016SoP}, CUB-200-2011~\cite{WahCUB_200_2011}), and
\textbf{(T,I)$\rightarrow$I} (FashionIQ~\cite{Wu2021FashionIQ}). These collectively
cover coarse-grained scene matching, fine-grained product/species discrimination,
and cross-modal composed retrieval. Detailed splits are in
Appendix~\ref{appendix:datasets}.

\paragraph{Implementation Details.}
For coarse retrieval (the standard retrieve-then-rerank prerequisite, not part
of our contribution), we use EVA-CLIP-18B~\cite{fang2023eva},
CLIP-ViT-B/32 and L/14~\cite{radford2021clip},
BLIP-2~\cite{li2023blip2}, and DINOv2~\cite{oquab2024dinov2} (for I$\rightarrow$I),
retrieving Top-$K$=20 candidates. For the MLLM teacher, we use
Gemini-3-flash~\cite{gemini3flash} and Gemini-3-pro~\cite{gemini3pro} with
temperature $T$=0.15. The student is Qwen3-VL-2B-Thinking~\cite{qwen3_vl_2b_thinking},
distilled with 20k queries (4K $\times$ 5 datasets). The distillation hyperparameters
$\lambda$ and $\eta$ are selected via grid search over $[0,1]$ with step 0.1 on
the validation set. We evaluate four variants: \textbf{EviRank-mini} (rubric-only,
no MLLM at test time), \textbf{EviRank} (distilled student),
\textbf{EviRank-plus} (rubric + Gemini-3-flash), and \textbf{EviRank-pro}
(rubric + Gemini-3-pro). Under the default $K{=}20$ and $M{=}5$, rubric
scoring runs \emph{locally} over all 20 candidates by reusing the
first-stage encoder and costs no MLLM call, so the online budget of
\textbf{EviRank-plus}/\textbf{-pro} is fixed at \emph{two} calls per query:
one evidence-extraction call (the \textsc{P-Cap}/\textsc{P-B0}/\textsc{P-B1}
modules of Appendix~\ref{appendix:prompts} are issued as a single structured
request) and one top-$M$ listwise call. All reported results and all
efficiency numbers use this same two-call configuration.
Full settings are in Appendix~\ref{appendix:impl_details}.

\paragraph{Baselines.}
We compare against AFS~\cite{khaertdinov2025little,Khaertdinov2025ALittleMore},
ImageScope~\cite{Luo2025ImageScope}, CoTRR~\cite{Wu2025CoTRR},
CoTMR~\cite{Sun2025CoTMR}, LoCoRE~\cite{Xiao2025LoCoRE}, ReMatch~\cite{rematch2025},
VLM2Vec~\cite{vlm2vec2025}, UniME-v2~\cite{unimev22026}, and
RetLLM~\cite{retllm2026}.

\begin{figure}[htpb]
  \centering
  \includegraphics[width=\columnwidth]{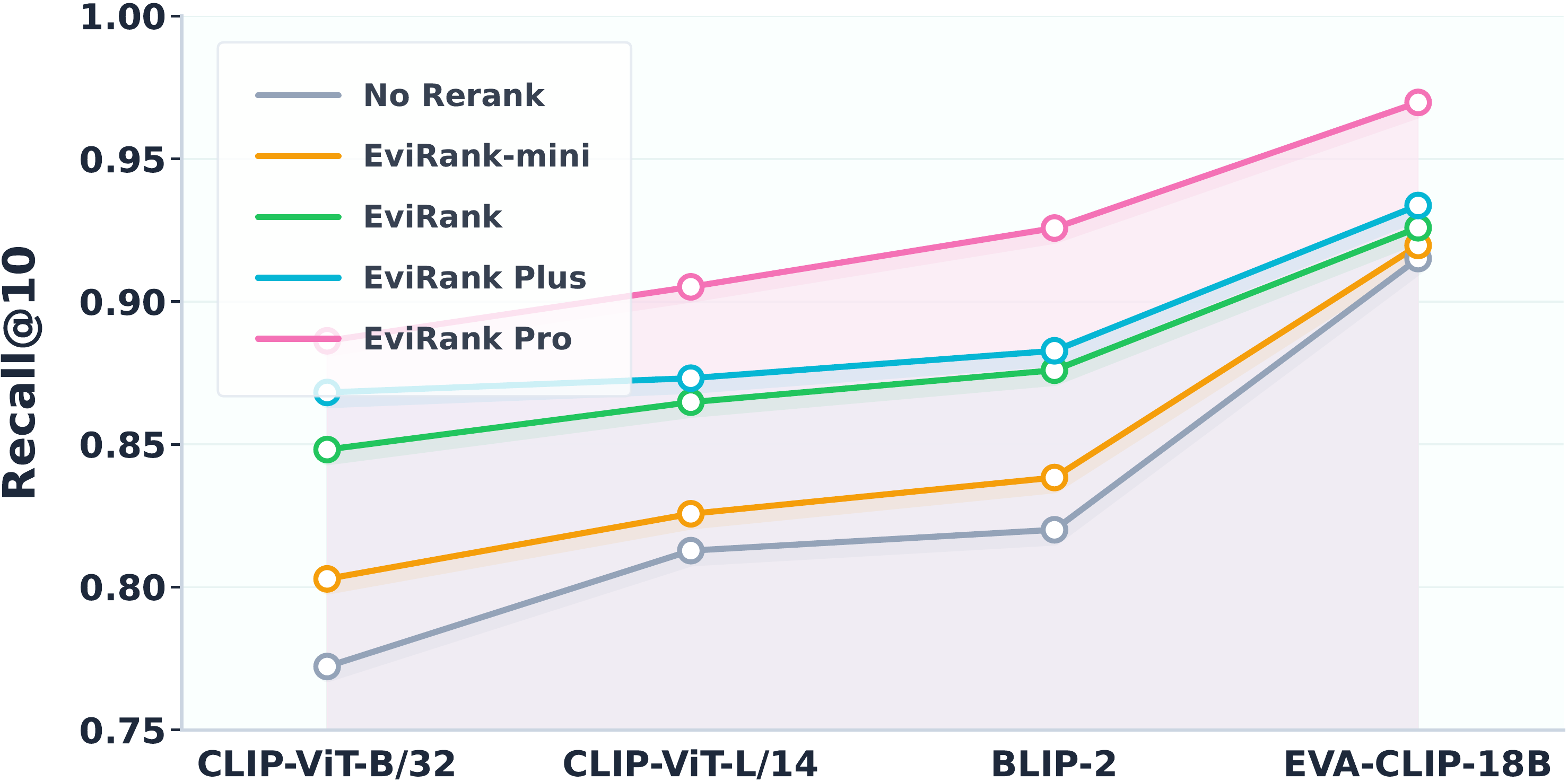}
  \caption{EviRank's Recall@10 on Flickr30k across the four
  coarse-retrieval backbones of Table~\ref{tab:flickr30k}.}
  \label{fig:flickr-compare}
\end{figure}

\subsection{Main Results}

\begin{table}[htpb]
\centering
\caption{Retrieval performance on COCO.}
\label{tab:coco}
\resizebox{\columnwidth}{!}{%
\fontsize{7.0}{8.0}\selectfont
\setlength{\tabcolsep}{3pt}
\renewcommand{\arraystretch}{0.90}
\begin{tabular}{
  L{0.75cm}
  L{1.4cm}
  S[table-format=2.1]
  S[table-format=2.1]
  S[table-format=2.1]
}
\toprule
\textbf{VLM} & \textbf{Methods}
& {\textbf{R@1}} & {\textbf{R@5}} & {\textbf{R@10}} \\
\midrule
& No ReRank    & 39.4 & 65.5 & 75.6 \\
& ImageScope   & 51.2 & 73.3 & 80.7 \\
& CoTRR        & 58.3 & 77.2 & 83.5 \\
\rowcolor{hl} & EviRank-mini & 59.9 & 77.4 & 82.4 \\
\rowcolor{hl} & EviRank      & 61.0 & 81.3 & 84.9 \\
\rowcolor{hl} & EviRank-plus & 62.3 & 83.7 & 86.6 \\
\rowcolor{hl}\multirow{-7}{*}{\rotatebox[origin=c]{90}{CLIP-ViT-B/32}}
& EviRank-pro  & 64.6 & 84.6 & 88.3 \\
\midrule
& No ReRank    & 46.4 & 71.1 & 79.7 \\
& ImageScope   & 53.7 & 75.9 & 83.5 \\
& CoTRR        & 59.9 & 79.7 & 84.7 \\
\rowcolor{hl} & EviRank-mini & 59.3 & 79.6 & 84.4 \\
\rowcolor{hl} & EviRank      & 61.7 & 81.7 & 85.9 \\
\rowcolor{hl} & EviRank-plus & 64.9 & 84.7 & 85.9 \\
\rowcolor{hl}\multirow{-7}{*}{\rotatebox[origin=c]{90}{CLIP-ViT-L/14}}
& EviRank-pro  & 69.5 & 87.7 & 89.4 \\
\bottomrule
\end{tabular}
}
\end{table}
Tables~\ref{tab:flickr30k}--\ref{tab:fashioniq},
Table~\ref{tab:newbaselines} (Appendix~\ref{appendix:newbaselines}),
and Figure~\ref{fig:flickr-compare}
demonstrate that \textbf{EviRank} consistently achieves SOTA across paradigms.
On Flickr30k with BLIP-2, \textbf{EviRank-pro} attains 95.61\% R@1, surpassing
CoTMR by 6.32 points; even rubric-only \textbf{EviRank-mini} outperforms all
prior methods without MLLM inference at test time. On COCO with CLIP-ViT-L/14,
\textbf{EviRank-pro} reaches 69.53\% R@1 (+9.6 over CoTRR). For I$\rightarrow$I,
\textbf{EviRank-pro} achieves 91.46\% (SoP) and 86.89\% (CUB-200) R@1, exceeding
LoCoRE-base by 7.66 and 8.59 points. On FashionIQ, R@10 improves up to 8.27 points
on Shirt over ImageScope. The gap between rubric-only \textbf{EviRank-mini} and
MLLM-enhanced variants validates that structured evidence and listwise reasoning
are \emph{complementary} rather than redundant.

\paragraph{Comparison with recent embedding-based re-rankers.}
Recent T$\rightarrow$I progress comes largely from stronger
\emph{representations} rather than from re-ranking, so we also
compare against four 2025--2026 systems of that kind
(Table~\ref{tab:newbaselines}): \textbf{EviRank-pro} reaches
95.6 / 69.5 R@1 on Flickr30k / COCO versus 85.6 / 62.8 for the
strongest of them (ReMatch). As these baselines already optimize
the embedding space that our first stage consumes, the gap comes
from evidence-conditioned verification, not from a better
retriever.

\begin{table}[htpb]
\centering
\caption{Retrieval performance (\%) on SoP and CUB-200 (DINOv2). mAP refers to mAP@R.}
\label{tab:sop_cub}
\resizebox{\columnwidth}{!}{%
\fontsize{7.0}{8.0}\selectfont
\setlength{\tabcolsep}{2pt}
\renewcommand{\arraystretch}{0.90}
\begin{tabular}{
  L{1.5cm}
  S[table-format=2.1] S[table-format=2.1] S[table-format=2.1] S[table-format=2.1]
  S[table-format=2.1] S[table-format=2.1] S[table-format=2.1] S[table-format=2.1]
}
\toprule
& \multicolumn{4}{c}{\textbf{SoP}} & \multicolumn{4}{c}{\textbf{CUB-200}} \\
\cmidrule(lr){2-5}\cmidrule(lr){6-9}
\textbf{Methods}
& {\textbf{R@1}} & {\textbf{R@5}} & {\textbf{R@10}} & {\textbf{mAP}}
& {\textbf{R@1}} & {\textbf{R@2}} & {\textbf{R@4}}  & {\textbf{mAP}} \\
\midrule
No ReRank    & 80.8 & 86.8 & 92.1 & 65.1 & 68.9 & 79.4 & 87.3 & 49.8 \\
SSR ReRank   & 81.2 & 87.4 & 91.9 & 66.7 & 69.4 & 79.0 & 86.1 & 54.2 \\
$\alpha$ QE  & 81.1 & 85.8 & 90.7 & 68.1 & 70.9 & 78.8 & 84.7 & 56.9 \\
RRT          & 81.9 & 86.7 & 92.4 & 67.2 & 68.7 & 85.0 & 95.6 & 55.6 \\
LoCoRE-tiny  & 82.4 & 87.8 & 93.1 & 68.0 & 71.4 & 86.8 & 96.4 & 58.1 \\
LoCoRE-small & 83.3 & 88.4 & 92.7 & 69.4 & 74.6 & 89.1 & 97.3 & 61.0 \\
LoCoRE-base  & 83.8 & 90.3 & 92.9 & 71.0 & 78.3 & 91.9 & 98.4 & 64.8 \\
\rowcolor{hl} EviRank-mini & 83.7 & 90.4 & 92.3 & 69.8 & 76.7 & 91.5 & 97.9 & 64.8 \\
\rowcolor{hl} EviRank      & 86.6 & 90.9 & 93.3 & 73.8 & 82.4 & 93.9 & 98.5 & 66.8 \\
\rowcolor{hl} EviRank-plus & 90.0 & 92.3 & 95.0 & 77.0 & 84.6 & 95.5 & 99.2 & 72.5 \\
\rowcolor{hl} EviRank-pro  & 91.5 & 93.9 & 96.0 & 79.6 & 86.9 & 97.6 & 99.4 & 76.6 \\
\bottomrule
\end{tabular}
}
\end{table}

\begin{table}[htpb]
\centering
\caption{Retrieval performance (\%) on FashionIQ.}
\label{tab:fashioniq}
\resizebox{\columnwidth}{!}{%
\fontsize{7.0}{8.0}\selectfont
\setlength{\tabcolsep}{2pt}
\renewcommand{\arraystretch}{0.90}
\begin{tabular}{
  L{0.75cm}
  L{1.5cm}
  S[table-format=2.1] S[table-format=2.1]
  S[table-format=2.1] S[table-format=2.1]
  S[table-format=2.1] S[table-format=2.1]
}
\toprule
\textbf{VLM} & \textbf{Methods}
& \multicolumn{2}{c}{\textbf{Shirt}}
& \multicolumn{2}{c}{\textbf{Dress}}
& \multicolumn{2}{c}{\textbf{Toptee}} \\
\cmidrule(lr){3-4}\cmidrule(lr){5-6}\cmidrule(lr){7-8}
& &
{\textbf{R@10}} & {\textbf{R@50}} &
{\textbf{R@10}} & {\textbf{R@50}} &
{\textbf{R@10}} & {\textbf{R@50}} \\
\midrule
& SEARLE       & 24.4 & 41.6 & 18.5 & 39.5 & 25.7 & 46.4 \\
& iSEARLE-OTI  & 27.0 & 43.4 & 21.2 & 42.1 & 26.8 & 48.7 \\
& CIReVL       & 28.3 & 47.8 & 25.2 & 46.3 & 31.2 & 53.8 \\
& ImageScope   & 31.6 & 37.4 & 26.8 & 46.3 & 35.8 & 55.9 \\
\rowcolor{hl} & EviRank-mini & 31.7 & 38.4 & 27.9 & 47.4 & 37.4 & 56.5 \\
\rowcolor{hl} & EviRank      & 33.8 & 40.5 & 32.0 & 49.7 & 42.9 & 59.0 \\
\rowcolor{hl} & EviRank-plus & 35.7 & 43.9 & 34.7 & 52.3 & 46.7 & 62.1 \\
\rowcolor{hl}\multirow{-8}{*}{\rotatebox[origin=c]{90}{CLIP-ViT-B/32}}
& EviRank-pro  & 39.7 & 46.7 & 36.8 & 56.7 & 49.0 & 64.8 \\
\midrule
& Pic2Word     & 26.2 & 43.6 & 20.0 & 40.2 & 27.9 & 47.4 \\
& iSEARLE-OTI  & 31.8 & 50.2 & 24.1 & 45.1 & 31.7 & 53.2 \\
& LDRE         & 31.0 & 51.2 & 22.9 & 46.7 & 31.5 & 53.6 \\
& FTI4CIR      & 31.3 & 50.5 & 24.4 & 47.8 & 32.4 & 54.2 \\
& ImageScope   & 32.8 & 51.0 & 26.1 & 46.1 & 35.0 & 55.1 \\
\rowcolor{hl} & EviRank-mini & 32.9 & 51.6 & 27.4 & 46.7 & 36.2 & 54.4 \\
\rowcolor{hl} & EviRank      & 35.0 & 54.8 & 30.6 & 50.0 & 39.8 & 57.6 \\
\rowcolor{hl} & EviRank-plus & 37.6 & 57.6 & 34.6 & 53.3 & 42.6 & 59.8 \\
\rowcolor{hl}\multirow{-9}{*}{\rotatebox[origin=c]{90}{CLIP-ViT-L/14}}
& EviRank-pro  & 40.1 & 59.9 & 38.9 & 56.8 & 49.7 & 64.6 \\
\bottomrule
\end{tabular}
}
\end{table}
Several older PRF/GRF/PALAVRA/SEARLE-OTI/iSEARLE/CIReVL/LinCIR baseline rows
are moved to Appendix~\ref{appendix:full_results}, which reproduces the full
tables.

\subsection{Ablation Studies}

\begin{table}[htpb]
\centering
\caption{Ablation results on T$\rightarrow$I and I$\rightarrow$I tasks with CLIP-ViT-L/14.}
\label{tab:ablation_ti_ii}
\newcommand{\drop}[2]{%
  #1\,{\tiny\textcolor{red}{$\downarrow$\!#2}}%
}
\resizebox{\columnwidth}{!}{%
\fontsize{7.5}{8.5}\selectfont
\setlength{\tabcolsep}{4pt}
\renewcommand{\arraystretch}{1.05}
\begin{tabular}{
  L{1.95cm}
  c c c c
}
\toprule
& \multicolumn{2}{c}{\textbf{T$\rightarrow$I}}
& \multicolumn{2}{c}{\textbf{I$\rightarrow$I}} \\
\cmidrule(lr){2-3}\cmidrule(lr){4-5}
\textbf{Variants}
& {\textbf{Flickr30K}} & {\textbf{COCO}}
& {\textbf{SoP}} & {\textbf{CUB-200}} \\
\midrule
\rowcolor{hl}
EviRank-pro
  & 96.9 & 89.4 & 87.6 & 99.4 \\
\midrule
w/o B0
  & \drop{94.5}{2.4\%} & \drop{88.6}{0.8\%}
  & \drop{82.3}{5.9\%} & \drop{96.5}{2.9\%} \\
w/o Slots
  & \drop{93.5}{3.4\%} & \drop{87.9}{1.6\%}
  & \drop{85.4}{2.4\%} & \drop{96.3}{3.1\%} \\
w/o Evidence
  & \drop{93.2}{3.8\%} & \drop{84.2}{5.7\%}
  & \drop{84.1}{3.9\%} & \drop{95.9}{3.5\%} \\
w/o Required
  & \drop{94.1}{2.8\%} & \drop{86.6}{3.0\%}
  & \drop{84.8}{3.0\%} & \drop{96.0}{3.4\%} \\
w/o Forbidden
  & \drop{94.6}{2.2\%} & \drop{87.4}{2.1\%}
  & \drop{85.6}{2.2\%} & \drop{97.4}{1.9\%} \\
w/o Ignore
  & \drop{96.4}{0.5\%} & \drop{88.8}{0.6\%}
  & \drop{87.3}{0.2\%} & \drop{99.4}{0.0\%} \\
\bottomrule
\end{tabular}
\label{tab:ablation1}
}
\end{table}

\begin{table}[htpb]
\centering
\caption{Ablation results on composed multimodal retrieval tasks with CLIP-ViT-L/14.}
\label{tab:ablation_tii}
\newcommand{\drop}[2]{%
  #1\,{\tiny\textcolor{red}{$\downarrow$\!#2}}%
}
\resizebox{\columnwidth}{!}{%
\fontsize{7.5}{8.5}\selectfont
\setlength{\tabcolsep}{5pt}
\renewcommand{\arraystretch}{1.05}
\begin{tabular}{
  L{1.95cm}
  c c c
}
\toprule
& \multicolumn{3}{c}{\textbf{(T,I)$\rightarrow$I}} \\
\cmidrule(lr){2-4}
\textbf{Variants}
& {\textbf{Shirt}} & {\textbf{Dress}} & {\textbf{Toptee}} \\
\midrule
\rowcolor{hl}
EviRank-pro
  & 40.1 & 38.9 & 49.7 \\
\midrule
w/o B0
  & \drop{37.6}{6.2\%} & \drop{35.4}{8.8\%}
  & \drop{46.5}{6.3\%} \\
w/o Slots
  & \drop{37.6}{6.0\%} & \drop{35.6}{8.2\%}
  & \drop{44.4}{10.5\%} \\
w/o Evidence
  & \drop{36.9}{7.8\%} & \drop{34.7}{10.5\%}
  & \drop{42.9}{13.5\%} \\
w/o Required
  & \drop{37.1}{7.3\%} & \drop{35.7}{8.0\%}
  & \drop{44.6}{10.0\%} \\
w/o Forbidden
  & \drop{37.9}{5.4\%} & \drop{37.3}{3.8\%}
  & \drop{46.8}{5.7\%} \\
w/o Ignore
  & \drop{39.2}{2.1\%} & \drop{38.4}{1.1\%}
  & \drop{48.6}{2.0\%} \\
\bottomrule
\end{tabular}
\label{tab:ablation2}
}
\end{table}
We ablate each component on all five benchmarks (CLIP-ViT-L/14 backbone, R@10
except CUB-200 R@4; Tables~\ref{tab:ablation1} and~\ref{tab:ablation2}).
Removing query understanding (w/o B0) most affects I$\rightarrow$I
(SoP $-$5.19, CUB $-$2.92), where caption generation bridges the modality gap.
Removing slot decomposition (w/o Slots) degrades performance everywhere,
especially on composed retrieval (Toptee $-$5.26), confirming that explicit
slot separation yields more systematic evidence than holistic descriptions,
and removing all evidence causes the largest drops (Toptee $-$6.71). Among
constraint types, Required anchors positive matching (COCO $-$2.73), Forbidden
disambiguates fine-grained candidates (CUB $-$1.96), and Ignore filters
non-discriminative variations (Toptee $-$1.04).

\begin{table}[htpb]
\centering
\small
\caption{Rubric vs.\ listwise component analysis.
M1: R@1 (T$\rightarrow$I/I$\rightarrow$I) / R@10 (CIR).
M2: R@10 / mAP / R@50 respectively.}
\label{tab:component}
\setlength{\tabcolsep}{4pt}
\begin{tabular}{lccc}
\toprule
\textbf{Setting} & \textbf{COCO} & \textbf{SoP} & \textbf{FashionIQ} \\
\midrule
Rubric-only      & 59.9 / 82.4 & 83.7 / 69.8 & 32.1 / 50.9 \\
Listwise-only    & 62.5 / 85.6 & 87.9 / 75.7 & 37.3 / 55.6 \\
\rowcolor{hl}\textbf{Rubric+Listwise} & \textbf{64.6 / 88.4} & \textbf{91.5 / 79.8} & \textbf{42.9 / 60.4} \\
\bottomrule
\end{tabular}
\end{table}

\subsection{Component Analysis: Rubric vs.\ Listwise}
\label{sec:component_analysis}
We further isolate \textbf{rubric scoring} versus \textbf{listwise
comparison} (Table~\ref{tab:component}). Each alone is strong, but their
combination is consistently best, with the largest gain on CIR (+5.6 R@10 on
FashionIQ over listwise-only): the rubric handles explicit constraints, while
listwise comparison captures implicit cues via joint visual comparison.

\subsection{Stability of Structured Evidence Extraction}
\label{sec:robustness}

A core claim of EviRank is that \emph{structuring} relevance
estimation yields stable, auditable signals rather than relying on
prompt wording. We validate it along two axes: stability of the
evidence schema, and gain attribution.

\paragraph{Stability of the evidence schema.}
We test stability of the structured-slot output along three axes
on FashionIQ (mean over 3 categories): (a) \emph{repeated runs}
(10$\times$ identical-input calls), (b) \emph{prompt
perturbation} (paraphrase, shuffle, typos), and
(c) \emph{cross-teacher} (Gemini-3-pro, GPT-5.4,
Claude-4.6-Opus/Sonnet, Gemini-3-flash). Across all settings,
Kendall's $\tau \geq 0.89$, Top-1 agreement $\geq 91\%$, and
R@10 std $\leq 1.3$. We attribute this stability to the fixed schema and ternary
R/F/I typing, which collapse the output space into a narrow,
JSON-checkable format across teachers. Stability is moreover
largely decoupled from teacher strength: replacing Gemini-3-pro
with the much cheaper Gemini-3-flash costs 4.6 R@10
(42.9$\rightarrow$38.3) yet keeps $\tau{=}0.90$ and 91.1\% Top-1
agreement---a weaker teacher orders candidates worse without
destabilizing \emph{what} evidence is extracted.
Detailed tables are in Appendix~\ref{appendix:robustness}.

\paragraph{Gains come from structured evidence, not prompt wording.}
\begin{table}[htpb]
\centering
\footnotesize
\caption{Prompt-engineering controlled experiments (R@1, \%).}
\label{tab:prompt_eng}
\setlength{\tabcolsep}{3pt}
\renewcommand{\arraystretch}{0.95}
\begin{adjustbox}{max width=\columnwidth}
\begin{tabular}{@{}lccc@{}}
\toprule
\textbf{Setting} & \textbf{COCO} & \textbf{SoP} & \textbf{FIQ R@10} \\
\midrule
Query-only listwise        & 63.0 & 84.7 & 36.6 \\
Free-form augmentation     & 63.5 & 86.7 & 38.3 \\
Caption only               & 64.8 & 85.3 & 37.9 \\
Evidence only              & 68.2 & 90.8 & 42.2 \\
\rowcolor{hl}\textbf{Caption + Evidence (Ours)} & \textbf{69.5} & \textbf{91.5} & \textbf{42.9} \\
\bottomrule
\end{tabular}
\end{adjustbox}
\end{table}

To check that improvements come from the slot-wise R/F/I
formulation rather than from prompt phrasing, we run three
controlled experiments (Table~\ref{tab:prompt_eng}; full results
in Appendix~\ref{appendix:prompt_sens}). Free-form augmentation
gives limited gains (+0.5 R@1 on COCO) while structured evidence
contributes most of the improvement (+6.5 R@1), and
prompt-template rewrites cause negligible variation
(R@1 std $<$0.7).

\subsection{Efficiency of Evidence-Conditioned Re-ranking}
\label{sec:efficiency}

For deployment, the distilled student inherits the
evidence-grounded ranking logic of Section~\ref{sec:distill}
without any teacher, evidence generator or runtime evidence cache.
On Flickr30k it takes $\sim$800ms per query while achieving
$\sim$10 R@1 points higher than CLIP coarse retrieval
($\sim$382ms); CLIP is the shared first stage of \emph{all}
re-ranking pipelines, so its latency is a fixed cost rather than
an alternative. The student is fully batchable: latency scales
linearly with $K$ and parallelizes across GPUs. With the teacher in the loop,
the default setting costs only two online calls per query
(Section~\ref{sec:setup}); for larger listwise batches,
divide-and-merge (Appendix~\ref{app:listwise_batch}) keeps total
calls linear in the number of groups and sequential depth
logarithmic.

\subsection{Case Study and Slot Semantics}

\begin{figure}[t]
  \centering
  \includegraphics[width=0.70\columnwidth]{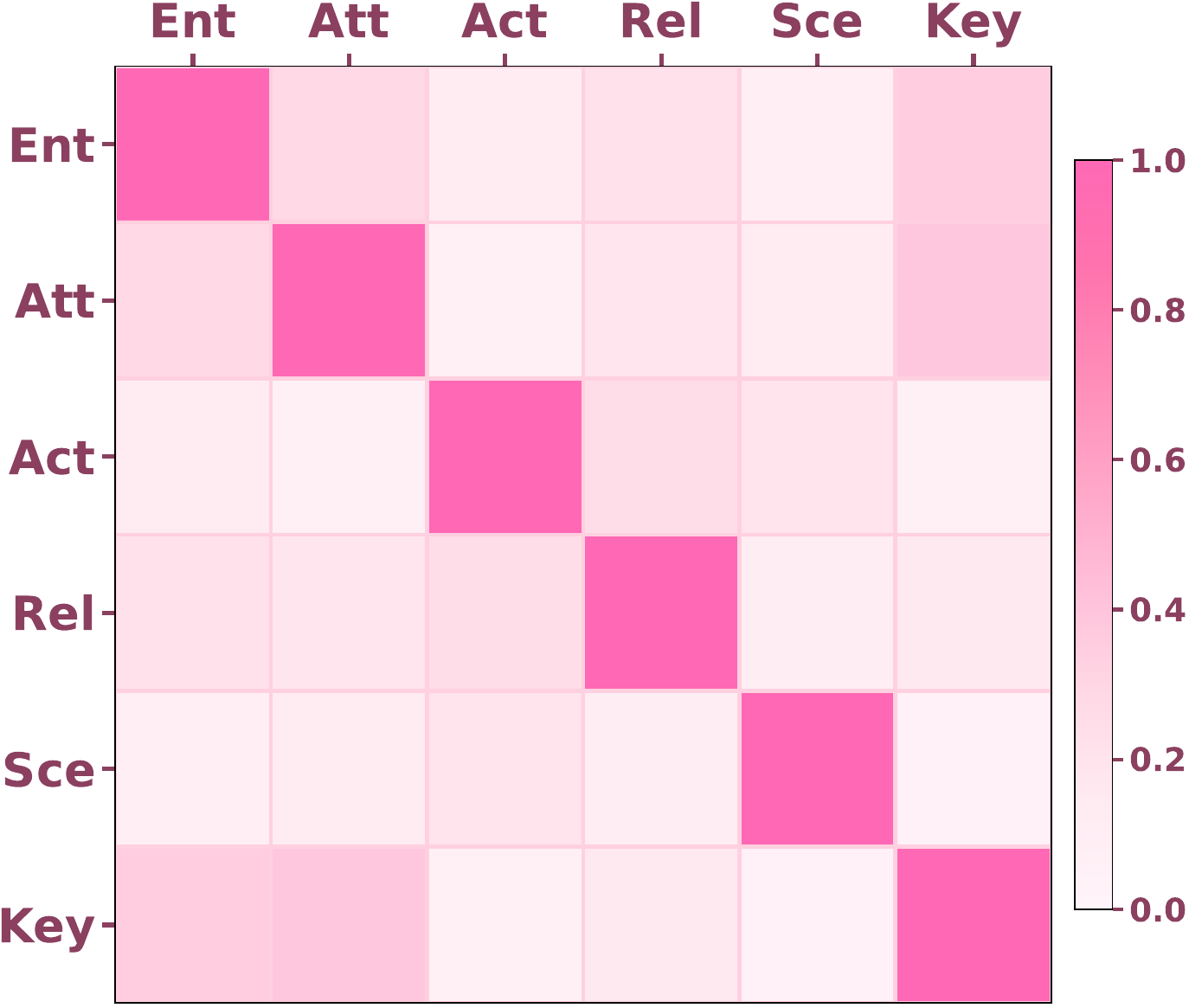}

  \caption{Pairwise SBERT similarity between the six evidence
  slots over 10k queries (off-diagonal mean 0.18).}
  \label{fig:heatmap1}
\end{figure}

\paragraph{Slots are near-orthogonal yet complementary.}
Figure~\ref{fig:heatmap1} shows pairwise SBERT similarity between
the six slots over 10k queries. The off-diagonal mean of 0.18
confirms they are not redundant restatements; the lowest pairs
separate distinct axes (\textit{Scene}--\textit{KeyDetails},
\textit{Actions}--\textit{Attributes}), while moderate
\textit{Entities}/\textit{Attributes}--\textit{KeyDetails}
correlations reflect a category-vs-instance hierarchy rather than
overlap.

\paragraph{Qualitative case.}
A bicycle query from SoP illustrates the pipeline:
\emph{required} slots anchor the correct category,
\emph{forbidden} slots reject the most confusable ones
(``mountain bike,'' ``road racing bicycle''), and
\emph{ignorable} slots absorb context noise (full case study and
R/F/I ablation in
Appendices~\ref{appendix:case_study}--\ref{appendix:slot_analysis}).

\section{Conclusion}

EviRank shows that multimodal image re-ranking is, at its core,
\emph{language-grounded semantic verification}: turning a query
into typed required / forbidden / ignorable constraints lets a
re-ranker reason about \emph{which} dimensions matter and
\emph{how} a candidate satisfies them, instead of committing the
decision to an opaque similarity score or an unstructured
chain-of-thought. Three findings recur across our five
benchmarks: rubric scoring and listwise comparison are
\emph{complementary}; \emph{forbidden} constraints, often
overlooked in retrieval design, are critical for fine-grained
discrimination; and structured evidence is robust across
teachers, prompts and decoding randomness. The same formulation
generalizes to NLP judgement tasks decomposable into typed atomic
units, and distils into a compact student retaining over 90\% of
the teacher's capability.

\section*{Acknowledgements}

This work was sponsored by the CCF-Tencent Rhino-Bird Open Research Fund
(No.~CCF-Tencent RAGR20250119), and was also supported by the Guangdong
Basic and Applied Basic Research Foundation (No.~2025A1515010304),
the Guangdong Province Project (No.~2024QN11X088), and the Guangzhou
Science and Technology Planning Project (No.~2025A03J4491).

\section*{Limitations}

Our evaluation uses widely adopted public image retrieval
benchmarks in English, which are standard for the community but do
not capture every deployment condition that real-world systems
encounter (e.g., multilingual queries, evolving content
distributions, or domain-specific corpora outside common benchmark
coverage). We also focus on still-image retrieval; extending the
framework to video, 3D or audio--visual content is an interesting
direction. Finally, we report retrieval metrics on offline
benchmarks; large-scale human-facing user studies in production
search are left as a natural next step.

\bibliography{main}

\clearpage
\appendix

\section{Listwise Batching and Rich Signal Package}
\label{app:listwise_batch}

This appendix complements Section~\ref{sec:verify} with the
batching strategy used for evidence-grounded listwise refinement,
and the full schema of the by-product supervision bundle that is
later reused in distillation (Section~\ref{sec:distill}).

\paragraph{Divide-and-merge batching.}
When the candidate pool is large ($K>8$), a single listwise call
becomes prompt-length-bound and unstable. We therefore partition
the candidates into groups of size $B$ (default $B{=}5$), rank
each group independently conditioned on the same evidence package
$\mathcal{E}(q)$, and iteratively merge the resulting sub-lists by
pairwise listwise re-ranking. Writing $n=\lceil K/B\rceil$ for the
number of groups, the procedure issues $n$ group calls plus $n-1$
pairwise merge calls, i.e.\ $2n-1=O(K/B)$ MLLM calls in total.
Because all merges at the same level of the binary merge tree are
independent, they can be issued concurrently, so the
\emph{sequential depth}---and hence wall-clock latency under a
parallel client---is $1+\lceil\log_2 n\rceil$ rounds. The
logarithmic factor therefore describes latency, not total call
count.

\paragraph{Rich signal package.}
A single listwise call produces, in addition to the permutation
$\pi=(\pi_1,\ldots,\pi_K)$, the following structured artifact for
each query--candidate list:
\begin{itemize}[leftmargin=*,nosep]
    \item Absolute relevance scores $\{r_i\}_{i=1}^{K}$ on a
    bounded scale (0--100), enabling calibrated comparisons across
    queries.
    \item A self-assessed confidence $p\in[0,1]$ reflecting how
    decisively the candidates can be ordered under the current
    evidence.
    \item A set of \emph{hard pairs} $\mathcal{H}=\{(i,j)\}$
    consisting of the closest competitors, each accompanied by a
    short distinguishing reason that references specific
    constraints in $\mathcal{E}(q)$.
    \item Optional slot-wise signals: per-slot binary indicators
    $y_{s,i}=\mathbf{1}[\text{Match}_s(c_i)>\theta]$ and the
    corresponding rates from the deterministic rubric.
\end{itemize}
All of these signals share a fixed JSON schema, are produced by
the same evidence-conditioned pipeline, and are consumed verbatim
by the distillation stage---no additional supervision labelling
is required.

\section{Additional Distillation Details}
\label{appendix:distill_loss}

The three component objectives and the confidence-weighted total
loss are defined in Section~\ref{sec:distill}
(Eq.~\ref{eq:total_loss}). This appendix records the motivation
behind each term and the implementation details omitted there.

\paragraph{Why teacher scores are standardized.}
The teacher reports absolute relevance on a bounded 0--100 scale,
but its effective dynamic range varies with query difficulty: an
unambiguous query may yield $\{95,90,20,\ldots\}$ while a hard one
yields $\{62,60,58,\ldots\}$. Per-query standardization by
$\mu_r,\sigma_r$ removes this query-level scale, so
$\mathcal{L}_{\text{score}}$ transfers the \emph{shape} of the
teacher's preference distribution rather than its absolute
calibration. Student scores are not standardized---they are
already produced by a trainable head---and both sides share the
single temperature $\tau$. The constant $\epsilon=10^{-8}$ only
guards against degenerate lists in which all teacher scores
coincide.

\paragraph{Why hard pairs are weighted by the inverse score gap.}
Hard pairs are exactly the comparisons that a bi-encoder gets
wrong, and their difficulty is not uniform: a pair separated by
one point of teacher score is far more informative than one
separated by thirty. The weight $w_{ij}$ therefore grows as the
gap $|r_i-r_j|$ shrinks, concentrating gradient mass on the
closest competitors, while the margin $\delta$ prevents the loss
from being driven to zero by an arbitrarily large score
difference. The pair set $\mathcal{H}$ is taken verbatim from the
teacher's listwise output; no additional mining is performed.

\paragraph{Slot auxiliary head.}
The slot head is a small linear projection on top of the
student's pooled query--candidate representation, producing one
logit per slot $s\in\mathcal{S}$, and is supervised only by the
teacher's binary indicators
$y_{s,i}=\mathbf{1}[\text{Match}_s(c_i)>\theta]$. It is used as a
regularizer that forces the shared representation to encode
constraint-level evidence rather than a single scalar preference;
the head is dropped after training, so deployed inference cost is
identical to a plain scoring student. When slot-wise labels are
missing for a query, $\mathcal{L}_{\text{slot}}$ is skipped for
that query rather than filled with pseudo-labels.

\paragraph{Hyperparameters and stability.}
We use $\tau{=}2.0$, $\delta{=}0.5$, $\lambda{=}0.5$ and
$\eta{=}0.3$, selected on the validation split; the full
configuration is listed in Table~\ref{tab:distill_hyper} in
Appendix~\ref{appendix:impl_details}. The confidence weight
$p\in(0,1)$ acts as a per-query learning-rate scaler, which we
found sufficient to keep training stable without additional
gradient clipping on low-confidence lists.

\section{Dataset Details}
\label{appendix:datasets}

We evaluate on five benchmarks spanning three retrieval paradigms.
Table~\ref{tab:dataset_summary} provides an overview, followed by
detailed descriptions of each dataset.

\begin{table}[h]
\centering
\caption{Summary of evaluation datasets.}
\label{tab:dataset_summary}
\small
\setlength{\tabcolsep}{3pt}
\begin{adjustbox}{max width=\columnwidth}
\begin{tabular}{@{}llrrr@{}}
\toprule
\textbf{Paradigm} & \textbf{Dataset} & \textbf{Images}
& \textbf{Classes} & \textbf{Test Split} \\
\midrule
\multirow{2}{*}{T$\rightarrow$I}
& MS~COCO        & 123,287 & --      & 5K \\
& Flickr30k      & 31,000  & --      & 1K \\
\midrule
\multirow{2}{*}{I$\rightarrow$I}
& SoP            & 120,053 & 22,634  & 60,502 \\
& CUB-200-2011   & 11,788  & 200     & 5,924 \\
\midrule
(T,I)$\rightarrow$I
& FashionIQ      & 77,684  & 3       & 6,016 \\
\bottomrule
\end{tabular}
\end{adjustbox}
\end{table}

\subsection{Text-to-Image Retrieval (T$\rightarrow$I)}

\noindent\textbf{MS~COCO}~\cite{Lin2014COCO} is a large-scale
image captioning and retrieval benchmark containing 123,287 images,
each annotated with five human-written captions. The dataset covers
a diverse range of everyday scenes with multiple objects and complex
spatial relationships. We follow the standard Karpathy
split~\cite{karpathy2015deep} and report results on the 5K test
set, where each text query is matched against all 5K images.

\noindent\textbf{Flickr30k}~\cite{Plummer2015Flickr30k} consists
of 31,000 images collected from the Flickr photo-sharing platform,
each accompanied by five descriptive captions. Compared to
MS~COCO, Flickr30k features more diverse photographic styles and
natural language descriptions with richer linguistic variation. We
evaluate on the commonly used 1K test split.

\subsection{Image-to-Image Retrieval (I$\rightarrow$I)}

\noindent\textbf{Stanford Online Products
(SoP)}~\cite{OhSong2016SoP} is an instance-level product retrieval
benchmark containing 120,053 images across 22,634 fine-grained
product categories sourced from eBay.com. The dataset is
characterized by large intra-class appearance variations (e.g.,
different viewpoints, lighting conditions, and backgrounds for the
same product) and high inter-class similarity among visually related
products. The standard split uses 59,551 images from 11,318
categories for training and 60,502 images from 11,316 categories
for testing.

\noindent\textbf{CUB-200-2011}~\cite{WahCUB_200_2011} is a
fine-grained visual recognition dataset consisting of 11,788 bird
images from 200 species. The dataset exhibits substantial
intra-class variations in pose, viewpoint, and appearance, while
inter-class differences can be extremely subtle (e.g.,
distinguishing between closely related warbler species). Following
the standard evaluation protocol, we use the first 100 species
(5,864 images) for training and the remaining 100 species (5,924
images) for testing.

\subsection{Composed Image Retrieval ((T,I)$\rightarrow$I)}

\noindent\textbf{FashionIQ}~\cite{Wu2021FashionIQ} is a benchmark
designed for image retrieval based on natural language feedback in
the fashion domain. It contains three categories---\emph{Dress},
\emph{Shirt}, and \emph{Toptee}---comprising 77,684 images and
30,134 image--text--image triplets. Each triplet consists of a
reference image, a target image, and two natural language
descriptions specifying how the target differs from the reference
(e.g., ``is more colorful and has shorter sleeves''). This setup
requires models to jointly reason over visual content and textual
modification intent. We report results on the standard validation
split following prior work~\cite{Luo2025ImageScope,Wu2025CoTRR}.

\section{Implementation Details}
\label{appendix:impl_details}

\subsection{Stage A: Coarse Retrieval}

We employ different backbone encoders depending on the retrieval
paradigm. For text-to-image retrieval on Flickr30k and MS~COCO, we
use EVA-CLIP-18B~\cite{fang2023eva},
CLIP-ViT-B/32~\cite{radford2021clip},
CLIP-ViT-L/14~\cite{radford2021clip}, and
BLIP-2~\cite{li2023blip2}. For image-to-image retrieval on SoP and
CUB-200-2011, we use DINOv2~\cite{oquab2024dinov2} following the
setup of LoCoRE~\cite{Xiao2025LoCoRE}. For composed image retrieval
on FashionIQ, we use CLIP-ViT-B/32 and CLIP-ViT-L/14. In all
settings, we retrieve top-$K$=20 candidates per query via cosine
similarity. The embedding-based matching operator $m_{\text{emb}}$
in the rubric scoring module reuses the same backbone encoder as
Stage~A.

\subsection{Stage B: Teacher Re-ranking}

We use Gemini-3-flash~\cite{gemini3flash} (for \textbf{EviRank-plus}) and
Gemini-3-pro~\cite{gemini3pro} (for \textbf{EviRank-pro}) as the multimodal
teacher models. All structured prompting (caption generation, query
expansion, evidence extraction, and listwise ranking) is performed
with temperature $T$=0.15 to encourage deterministic outputs. For
listwise re-ranking, candidates are processed in batches of 5; when
$K > 8$, a divide-and-merge strategy is applied as described in
Appendix~\ref{app:listwise_batch}. The rubric scoring hyperparameters in
Eq.~\ref{eq:rubric} are configured as follows: slot weights $w_s$ are set
uniformly to $1/|\mathcal{S}|$ (i.e., $1/6$), the forbidden penalty
$\beta$=0.75, and the consistency weight $\gamma$=0.1.

\subsection{Stage C: Knowledge Distillation}

We construct the distillation dataset by uniformly sampling 20k
queries across all five datasets (4K per dataset: Flickr30k,
MS~COCO, SoP, CUB-200-2011, and FashionIQ). For each sampled
query, we execute the full Stage~B pipeline using Gemini-3-pro as
the teacher to produce the complete supervision package, including
listwise rankings, absolute relevance scores (0--100), confidence
estimates, hard-pair annotations, and per-slot hit/violation rates.
Candidate ordering is randomly shuffled before teacher inference to
eliminate position bias.

The student model is initialized from
Qwen3-VL-2B-Thinking~\cite{qwen3_vl_2b_thinking} and fine-tuned
for 10 epochs using AdamW with cosine learning rate scheduling.
Table~\ref{tab:distill_hyper} summarizes the full hyperparameter
configuration.

\begin{table}[h]
\centering
\caption{Distillation hyperparameters for Stage~C.}
\label{tab:distill_hyper}
\small
\begin{tabular}{ll}
\toprule
\textbf{Hyperparameter} & \textbf{Value} \\
\midrule
Student backbone        & Qwen3-VL-2B-Thinking \\
Training epochs         & 10 \\
Batch size              & 32 \\
Learning rate           & $5 \times 10^{-5}$ \\
Optimizer               & AdamW \\
LR schedule             & Cosine decay \\
KL temperature $\tau$   & 2.0 \\
Hard-pair margin $\delta$ & 0.5 \\
Pair loss weight $\lambda$ & 0.5 \\
Slot loss weight $\eta$ & 0.3 \\
Distillation queries    & 20k (4K $\times$ 5 datasets) \\
Teacher model           & Gemini-3-pro \\
Hardware                & 8 $\times$ NVIDIA H20 80GB \\
\bottomrule
\end{tabular}
\end{table}

\subsection{Model Variants}

Table~\ref{tab:variants} summarizes the four \textbf{EviRank} variants and
their corresponding configurations.

\begin{table}[h]
\centering
\caption{Summary of EviRank variants.}
\label{tab:variants}
\small
\setlength{\tabcolsep}{3pt}
\begin{adjustbox}{max width=\columnwidth}
\begin{tabular}{@{}lccl@{}}
\toprule
\textbf{Variant} & \textbf{Rubric} & \textbf{Listwise}
& \textbf{Inference Model} \\
\midrule
EviRank-mini & $m_{\text{emb}}$ & \xmark
& CLIP / DINOv2 only \\
EviRank      & $m_{\text{emb}}$ & \xmark
& Qwen3-VL-2B (distilled) \\
EviRank-plus & $m_{\text{emb}}$ & \cmark
& Gemini-3-flash \\
EviRank-pro  & $m_{\text{emb}}$ & \cmark
& Gemini-3-pro \\
\bottomrule
\end{tabular}
\end{adjustbox}
\end{table}

\section{Comparison with Recent T$\rightarrow$I Baselines}
\label{appendix:newbaselines}

\begin{table}[htpb]
\centering
\small
\caption{Comparison with recent SOTA T$\rightarrow$I baselines (R@1, \%).}
\label{tab:newbaselines}
\setlength{\tabcolsep}{6pt}
\begin{tabular}{lcc}
\toprule
\textbf{Method} & \textbf{Flickr30k} & \textbf{COCO} \\
\midrule
VLM2Vec      & 80.0 & 49.2 \\
RetLLM       & 82.0 & 54.1 \\
UniME-v2     & 85.5 & 60.9 \\
ReMatch     & 85.6 & 62.8 \\
\rowcolor{hl}\textbf{EviRank-pro} & \textbf{95.6} & \textbf{69.5} \\
\bottomrule
\end{tabular}
\end{table}

Table~\ref{tab:newbaselines} adds a head-to-head comparison with very recent
embedding/CoT-based methods. \textbf{EviRank} substantially outperforms all of
them on both benchmarks.

\section{Full Main-Result Tables}
\label{appendix:full_results}

For space, the main-text tables (Tables~\ref{tab:flickr30k}--\ref{tab:fashioniq})
omit several older CIR/CIR-style baselines (PRF, GRF, PALAVRA, SEARLE-OTI,
iSEARLE, CIReVL, LinCIR, AQE, DQE, etc.). For completeness and reproducibility,
we reproduce the full tables below
(Tables~\ref{tab:flickr30k_full}, \ref{tab:sop_cub_full}
and~\ref{tab:fashioniq_full}).

\begin{table}[h]
\centering
\caption{Full Flickr30k results across all backbones and baselines.}
\label{tab:flickr30k_full}
\resizebox{\columnwidth}{!}{%
\fontsize{7.0}{8.0}\selectfont
\setlength{\tabcolsep}{3pt}
\renewcommand{\arraystretch}{0.92}
\begin{tabular}{
  L{1.2cm}
  L{1.8cm}
  S[table-format=2.2]
  S[table-format=2.2]
  S[table-format=2.2]
  S[table-format=2.2]
}
\toprule
\textbf{Backbone} & \textbf{Methods}
& {\textbf{R@1}} & {\textbf{R@5}} & {\textbf{R@10}} & {\textbf{MRR@5}} \\
\midrule
\multirow{9}{*}{\rotatebox[origin=c]{90}{EVA-CLIP-18B}}
& No ReRank    & 84.08 & 95.33 & 96.97 & 89.75 \\
& ImageScope   & 84.89 & 95.68 & 97.01 & 90.12 \\
& CoTRR        & 85.69 & 96.21 & 97.34 & 90.37 \\
& CoTMR        & 85.93 & 96.82 & 97.69 & 90.96 \\
& EviRank-mini & 84.89 & 96.47 & 97.52 & 90.97 \\
& EviRank      & 86.65 & 96.93 & 98.01 & 91.11 \\
& EviRank-plus & 87.19 & 97.80 & 98.41 & 92.20 \\
& EviRank-pro  & 87.98 & 97.65 & 98.73 & 92.51 \\
\midrule
\multirow{12}{*}{\rotatebox[origin=c]{90}{CLIP-ViT-B/32}}
& No ReRank    & 67.10 & 89.00 & 90.03 & 75.80 \\
& PRF          & 66.90 & 89.20 & 90.23 & 75.70 \\
& GRF          & 71.60 & 89.60 & 90.42 & 78.90 \\
& AFS          & 72.40 & 91.30 & 91.89 & 80.10 \\
& Explicit     & 72.50 & 93.70 & 93.96 & 80.90 \\
& ImageScope   & 78.84 & 92.66 & 95.64 & 80.23 \\
& CoTRR        & 84.68 & 94.72 & 96.08 & 80.99 \\
& CoTMR        & 84.77 & 94.89 & 96.39 & 81.18 \\
& EviRank-mini & 85.19 & 93.97 & 96.42 & 82.67 \\
& EviRank      & 86.18 & 95.03 & 96.74 & 86.78 \\
& EviRank-plus & 87.12 & 96.31 & 97.09 & 89.91 \\
& EviRank-pro  & 87.24 & 97.17 & 97.93 & 90.19 \\
\midrule
\multirow{12}{*}{\rotatebox[origin=c]{90}{CLIP-ViT-L/14}}
& No ReRank    & 72.70 & 91.20 & 92.45 & 80.00 \\
& PRF          & 72.60 & 91.70 & 92.63 & 80.00 \\
& GRF          & 76.60 & 92.80 & 93.12 & 83.30 \\
& AFS          & 78.40 & 94.30 & 95.20 & 84.60 \\
& Explicit     & 74.60 & 93.30 & 94.26 & 82.00 \\
& ImageScope   & 78.10 & 91.01 & 93.38 & 82.26 \\
& CoTRR        & 83.81 & 93.23 & 94.27 & 82.89 \\
& CoTMR        & 84.52 & 93.97 & 94.39 & 83.03 \\
& EviRank-mini & 83.62 & 92.41 & 93.37 & 82.29 \\
& EviRank      & 84.97 & 94.33 & 95.39 & 84.93 \\
& EviRank-plus & 86.74 & 94.63 & 96.10 & 85.43 \\
& EviRank-pro  & 88.69 & 95.48 & 96.91 & 87.06 \\
\midrule
\multirow{12}{*}{\rotatebox[origin=c]{90}{BLIP-2}}
& No ReRank    & 86.80 & 97.50 & 98.02 & 91.50 \\
& PRF          & 86.90 & 97.60 & 98.13 & 91.70 \\
& GRF          & 89.30 & 98.50 & 98.69 & 93.20 \\
& AFS          & 89.00 & 98.20 & 98.30 & 93.10 \\
& Explicit     & 88.60 & 98.40 & 98.56 & 92.80 \\
& ImageScope   & 88.81 & 98.27 & 98.38 & 92.89 \\
& CoTRR        & 88.93 & 98.31 & 98.42 & 93.08 \\
& CoTMR        & 89.29 & 98.49 & 98.90 & 93.42 \\
& EviRank-mini & 89.33 & 98.57 & 98.89 & 93.27 \\
& EviRank      & 91.34 & 98.62 & 99.01 & 93.57 \\
& EviRank-plus & 93.52 & 99.03 & 99.23 & 94.21 \\
& EviRank-pro  & 95.61 & 99.22 & 99.41 & 95.62 \\
\bottomrule
\end{tabular}
}
\end{table}

\begin{table}[h]
\centering
\caption{Full SoP and CUB-200 results with all I$\rightarrow$I baselines.}
\label{tab:sop_cub_full}
\resizebox{\columnwidth}{!}{%
\fontsize{7.0}{8.0}\selectfont
\setlength{\tabcolsep}{3pt}
\renewcommand{\arraystretch}{0.92}
\begin{tabular}{
  L{1.95cm}
  S[table-format=2.2] S[table-format=2.2] S[table-format=2.2] S[table-format=2.2]
  S[table-format=2.2] S[table-format=2.2] S[table-format=2.2] S[table-format=2.2]
}
\toprule
& \multicolumn{4}{c}{\textbf{SoP}} & \multicolumn{4}{c}{\textbf{CUB-200}} \\
\cmidrule(lr){2-5}\cmidrule(lr){6-9}
\textbf{Methods}
& {\textbf{R@1}} & {\textbf{R@5}} & {\textbf{R@10}} & {\textbf{mAP}}
& {\textbf{R@1}} & {\textbf{R@2}} & {\textbf{R@4}}  & {\textbf{mAP}} \\
\midrule
No ReRank    & 80.80 & 86.87 & 92.10 & 65.10 & 68.90 & 79.40 & 87.30 & 49.80 \\
SSR ReRank   & 81.20 & 87.42 & 91.90 & 66.70 & 69.40 & 79.00 & 86.10 & 54.20 \\
AQE          & 76.90 & 73.69 & 89.30 & 66.10 & 66.90 & 76.80 & 82.70 & 58.40 \\
DQE          & 67.90 & 72.71 & 81.50 & 47.80 & 67.00 & 75.50 & 82.00 & 54.60 \\
$\alpha$ QE  & 81.10 & 85.86 & 90.70 & 68.10 & 70.90 & 78.80 & 84.70 & 56.90 \\
RRT          & 81.90 & 86.74 & 92.40 & 67.20 & 68.70 & 85.00 & 95.60 & 55.60 \\
LoCoRE-tiny  & 82.40 & 87.80 & 93.10 & 68.00 & 71.40 & 86.80 & 96.40 & 58.10 \\
LoCoRE-small & 83.30 & 88.49 & 92.70 & 69.40 & 74.60 & 89.10 & 97.30 & 61.00 \\
LoCoRE-base  & 83.80 & 90.37 & 92.90 & 71.00 & 78.30 & 91.90 & 98.40 & 64.80 \\
EviRank-mini & 83.69 & 90.41 & 92.34 & 69.83 & 76.72 & 91.49 & 97.90 & 64.78 \\
EviRank      & 86.56 & 90.88 & 93.27 & 73.80 & 82.38 & 93.94 & 98.48 & 66.79 \\
EviRank-plus & 89.99 & 92.33 & 94.97 & 77.00 & 84.56 & 95.48 & 99.23 & 72.48 \\
EviRank-pro  & 91.46 & 93.93 & 96.02 & 79.58 & 86.89 & 97.62 & 99.44 & 76.63 \\
\bottomrule
\end{tabular}
}
\end{table}

\begin{table}[h]
\centering
\caption{Full FashionIQ results across all baselines and backbones.}
\label{tab:fashioniq_full}
\resizebox{\columnwidth}{!}{%
\fontsize{7.0}{8.0}\selectfont
\setlength{\tabcolsep}{3pt}
\renewcommand{\arraystretch}{0.92}
\begin{tabular}{
  L{0.85cm}
  L{2.05cm}
  S[table-format=2.2] S[table-format=2.2]
  S[table-format=2.2] S[table-format=2.2]
  S[table-format=2.2] S[table-format=2.2]
}
\toprule
\textbf{VLM} & \textbf{Methods}
& \multicolumn{2}{c}{\textbf{Shirt}}
& \multicolumn{2}{c}{\textbf{Dress}}
& \multicolumn{2}{c}{\textbf{Toptee}} \\
\cmidrule(lr){3-4}\cmidrule(lr){5-6}\cmidrule(lr){7-8}
& &
{\textbf{R@10}} & {\textbf{R@50}} &
{\textbf{R@10}} & {\textbf{R@50}} &
{\textbf{R@10}} & {\textbf{R@50}} \\
\midrule
\multirow{12}{*}{\rotatebox[origin=c]{90}{CLIP-ViT-B/32}}
& PALAVRA      & 21.49 & 37.05 & 17.25 & 35.94 & 20.55 & 38.76 \\
& SEARLE       & 24.44 & 41.61 & 18.54 & 39.51 & 25.70 & 46.46 \\
& SEARLE-OTI   & 25.37 & 41.32 & 17.85 & 39.91 & 24.12 & 45.79 \\
& iSEARLE      & 25.81 & 43.52 & 20.92 & 42.19 & 26.47 & 48.70 \\
& iSEARLE-OTI  & 27.09 & 43.42 & 21.27 & 42.19 & 26.82 & 48.75 \\
& CIReVL       & 28.36 & 47.84 & 25.29 & 46.36 & 31.21 & 53.85 \\
& LDRE         & 27.38 & 46.27 & 19.97 & 41.84 & 27.07 & 48.78 \\
& ImageScope   & 31.65 & 37.49 & 26.82 & 46.31 & 35.80 & 55.94 \\
& EviRank-mini & 31.73 & 38.42 & 27.93 & 47.39 & 37.43 & 56.52 \\
& EviRank      & 33.79 & 40.45 & 31.96 & 49.73 & 42.89 & 58.98 \\
& EviRank-plus & 35.65 & 43.87 & 34.73 & 52.33 & 46.74 & 62.13 \\
& EviRank-pro  & 39.67 & 46.66 & 36.80 & 56.65 & 48.98 & 64.79 \\
\midrule
\multirow{14}{*}{\rotatebox[origin=c]{90}{CLIP-ViT-L/14}}
& Pic2Word     & 26.20 & 43.60 & 20.00 & 40.20 & 27.90 & 47.40 \\
& SEARLE       & 26.89 & 45.58 & 20.48 & 43.13 & 29.32 & 49.97 \\
& SEARLE-OTI   & 30.37 & 47.49 & 21.57 & 44.47 & 30.90 & 51.76 \\
& iSEARLE      & 28.75 & 47.84 & 22.51 & 46.36 & 31.31 & 52.68 \\
& iSEARLE-OTI  & 31.80 & 50.20 & 24.19 & 45.12 & 31.72 & 53.29 \\
& CIReVL       & 29.49 & 47.40 & 24.79 & 44.76 & 31.36 & 53.65 \\
& LDRE         & 31.04 & 51.22 & 22.93 & 46.76 & 31.57 & 53.64 \\
& LinCIR       & 29.10 & 46.81 & 20.92 & 42.44 & 28.81 & 50.18 \\
& FTI4CIR      & 31.35 & 50.59 & 24.49 & 47.84 & 32.43 & 54.21 \\
& ImageScope   & 32.87 & 51.07 & 26.17 & 46.15 & 35.03 & 55.12 \\
& EviRank-mini & 32.85 & 51.58 & 27.36 & 46.74 & 36.18 & 54.43 \\
& EviRank      & 34.96 & 54.83 & 30.64 & 49.96 & 39.83 & 57.61 \\
& EviRank-plus & 37.63 & 57.62 & 34.56 & 53.31 & 42.57 & 59.84 \\
& EviRank-pro  & 40.12 & 59.89 & 38.89 & 56.75 & 49.68 & 64.59 \\
\bottomrule
\end{tabular}
}
\end{table}

\section{Complete Case Study: Evidence-Based Bicycle Retrieval}
\label{appendix:case_study}

\begin{figure}[h]
  \centering
  \includegraphics[width=\columnwidth]{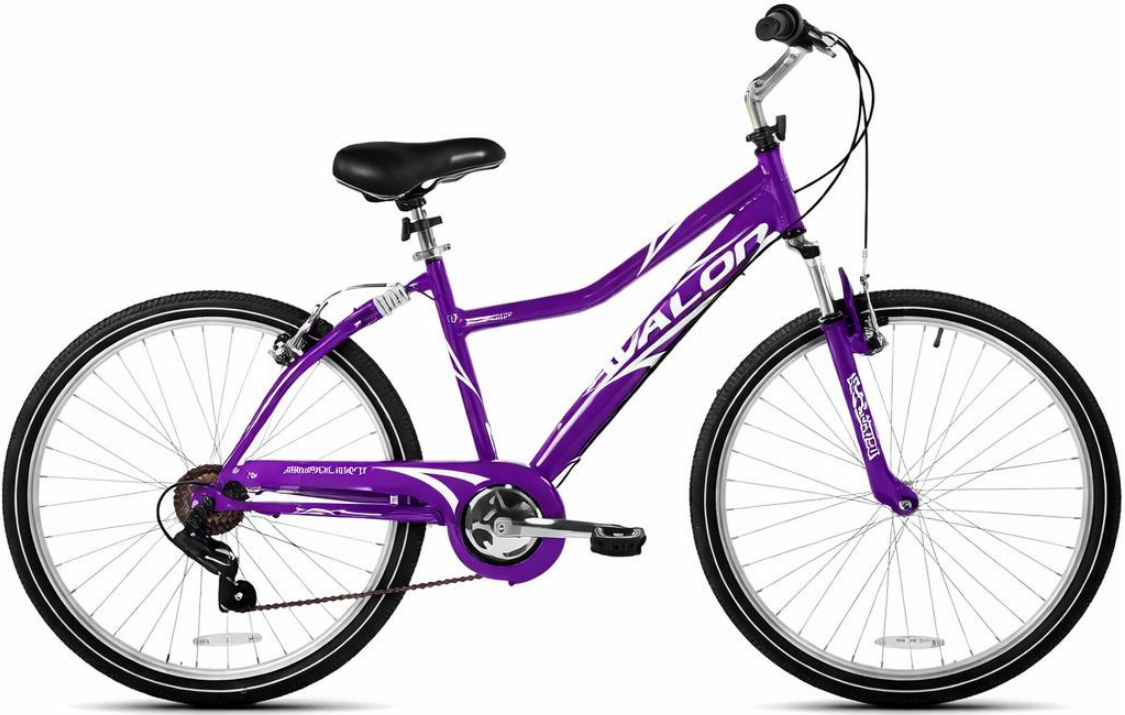}
  \caption{Case study's bicycle case.}
  \label{fig:case_study_appendix}
\end{figure}
This appendix provides comprehensive analysis of \textbf{EviRank}'s evidence extraction for the bicycle case study shown in Figure~\ref{fig:case_study_appendix}. Beyond presenting the extracted evidence, we analyze \textit{why} each constraint emerges and \textit{how} it contributes to retrieval accuracy.

\subsection{Query Processing Pipeline}

\subsubsection{Stage 1: Caption Generation}

\textbf{Generated Caption:}
\begin{quote}
\textit{This is a women's leisure bicycle featuring a low step-through curved frame design. The overall silhouette is sleek and streamlined, oriented toward daily commuting and light recreational use. The bicycle is primarily finished in a bright purple metallic color with glossy paint, complemented by black saddle, handlebars, and tires. Structural components include spoked wheel assemblies, upright handlebars, a comfort saddle, and basic drivetrain and braking systems. A prominent brand logo is printed on the frame side.}
\end{quote}

\textbf{Analysis:} The initial caption establishes three critical semantic anchors: (1) \textit{product category} (women's leisure bicycle), (2) \textit{distinguishing structure} (low step-through frame), and (3) \textit{primary visual attribute} (bright purple metallic). These anchors are essential because bicycle retrieval failures typically occur when the system confuses leisure bikes with mountain bikes or road bikes—categories that share similar components but differ fundamentally in frame geometry and intended use.

\subsubsection{Stage 2: Query Expansion}

\textbf{Expanded Description:}
\begin{quote}
\textit{A leisure/commuter women's bicycle with a classic low step-through curved frame configuration, smooth modern lines, and a style balancing sporty aesthetics with everyday practicality. Primary color is bright purple, accented with white brand lettering and minimal black components. Metal frame and fork with smooth glossy paint finish. Full-size spoked wheel assemblies, stable frame geometry from curved top and down tubes, widened comfort saddle, upright handlebars, and rear single-speed or simple derailleur system. Elongated proportions emphasizing riding stability. Brand logos on frame side, color-matched chain guard and fender areas, functional pedals, chain, braking system, and adjustable seat post—suitable for daily riding and light outdoor use.}
\end{quote}

\textbf{Analysis:} The expansion serves two purposes. First, it adds \textit{discriminative details} that were implicit in the image: ``elongated proportions,'' ``comfort saddle,'' and ``upright handlebars'' collectively define the leisure category's ergonomic signature. Second, it introduces \textit{component-level vocabulary} (chain guard, fender, derailleur) that enables precise matching against product descriptions in the gallery. Without this expansion, the retrieval system would rely solely on global visual similarity, missing the fine-grained attributes that distinguish near-identical products.

\subsection{Evidence Extraction with Dimensional Analysis}

\textbf{Global Summary:}
\begin{quote}
\textit{The image displays a women's leisure/commuter bicycle predominantly finished in bright purple, featuring a low step-through curved frame with elongated and flowing lines. The configuration emphasizes stability and comfort for daily riding, suitable for urban commuting and light outdoor use.}
\end{quote}

\subsubsection{Entities: Category-Level Disambiguation}

\textbf{Required:}
\begin{itemize}[leftmargin=*,nosep]
    \item \textit{The image features a women's leisure/commuter bicycle as the main subject.}
    \item \textit{This product is a modern-style city bike with a low step-through frame structure and upright riding orientation.}
\end{itemize}

\textbf{Forbidden:}
\begin{itemize}[leftmargin=*,nosep]
    \item \textit{The image features a mountain bike as the main subject, characterized by robust suspension structure and off-road orientation.}
    \item \textit{This product is a road racing bicycle, featuring a high top-tube frame and extremely narrow tire configuration.}
\end{itemize}

\textbf{Ignore:}
\begin{itemize}[leftmargin=*,nosep]
    \item \textit{The image contains simple and clean presentation environment elements.}
    \item \textit{The frame may contain minor photographic cropping variations.}
\end{itemize}

\textbf{Why These Constraints Emerge:} Entity-level evidence targets the most consequential retrieval errors—\textit{category confusion}. In embedding space, a purple leisure bike may be closer to a purple mountain bike than to a green leisure bike, because color dominates the learned representation. The required constraints anchor the search to the correct product category, while forbidden constraints explicitly reject the nearest confusable categories.

\textbf{Functional Role:} During re-ranking, candidates satisfying entity requirements receive positive scores, while those matching forbidden descriptions receive strong penalties. This creates a ``category gate'' that filters out high-similarity but wrong-category results before finer attributes are evaluated.

\subsubsection{Attributes: Visual Property Specification}

\textbf{Required:}
\begin{itemize}[leftmargin=*,nosep]
    \item \textit{The product's primary color is bright purple, complemented by white brand lettering and minimal black components.}
    \item \textit{The frame and fork are constructed from metal material, presenting a smooth glossy paint texture.}
\end{itemize}

\textbf{Forbidden:}
\begin{itemize}[leftmargin=*,nosep]
    \item \textit{The product's primary color is red or orange, complemented by large-area dark camouflage patterns.}
    \item \textit{The frame material is carbon fiber or plastic, presenting a matte or rough surface texture.}
\end{itemize}

\textbf{Ignore:}
\begin{itemize}[leftmargin=*,nosep]
    \item \textit{Minor shade variations exist within the same purple color family.}
    \item \textit{Normal reflection differences exist on local metal surfaces.}
\end{itemize}

\textbf{Why These Constraints Emerge:} Attribute evidence addresses \textit{intra-category discrimination}. Once the category is established, retrieval must distinguish among dozens of similar leisure bicycles. Color (bright purple vs.\ red/orange) and material finish (glossy metal vs.\ matte carbon) are the primary differentiators at this level.

\textbf{Functional Role:} The ignore constraints are particularly important here. Without them, lighting-induced color shifts or reflection variations would cause false negatives—rejecting correct matches due to superficial differences. By explicitly modeling acceptable variation, \textbf{EviRank} achieves robustness that rigid attribute matching cannot provide.

\subsubsection{Actions: State and Presentation Mode}

\textbf{Required:}
\begin{itemize}[leftmargin=*,nosep]
    \item \textit{The bicycle is displayed completely from a side view angle, with the body maintaining an upright balanced state.}
    \item \textit{The vehicle is in a fully assembled state ready for immediate riding, with all functional components installed.}
\end{itemize}

\textbf{Forbidden:}
\begin{itemize}[leftmargin=*,nosep]
    \item \textit{The bicycle is displayed in a folded state, with the frame and wheels collapsed together.}
    \item \textit{The vehicle is in a disassembled or maintenance state, with multiple key components placed separately.}
\end{itemize}

\textbf{Ignore:}
\begin{itemize}[leftmargin=*,nosep]
    \item \textit{Minor forward or backward offset exists in the shooting angle.}
    \item \textit{Small variations exist in wheel orientation during display.}
\end{itemize}

\textbf{Why These Constraints Emerge:} Product images of the same bicycle may appear in vastly different states—assembled, partially disassembled, or folded (for folding bikes). The action dimension ensures that presentation state does not confuse identity matching.

\textbf{Functional Role:} This dimension prevents a subtle failure mode: matching a fully assembled leisure bike to component images or maintenance photos of the same product line. The ignore constraints allow matching across minor viewpoint variations that are common in e-commerce photography.

\subsubsection{Relations: Structural Geometry Verification}

\textbf{Required:}
\begin{itemize}[leftmargin=*,nosep]
    \item \textit{The front and rear wheels are correctly connected to the frame's front fork and rear dropout through spoked wheel assemblies.}
    \item \textit{The handlebars, seat post, and frame form an upright comfortable geometric relationship, emphasizing stable riding posture.}
\end{itemize}

\textbf{Forbidden:}
\begin{itemize}[leftmargin=*,nosep]
    \item \textit{The front and rear wheels exhibit obvious misalignment with the frame, with abnormal structural proportions.}
    \item \textit{The handlebars and frame are connected through complex linkage structures, forming an unconventional layout.}
\end{itemize}

\textbf{Ignore:}
\begin{itemize}[leftmargin=*,nosep]
    \item \textit{Minor spacing differences exist between components.}
    \item \textit{Local structural alignment variations exist within normal tolerance ranges.}
\end{itemize}

\textbf{Why These Constraints Emerge:} Relational evidence captures the \textit{geometric signature} of a product category. Leisure bicycles have a characteristic relationship between handlebars, seat, and frame that differs from racing bikes (aggressive forward lean) or mountain bikes (suspension-mediated geometry).

\textbf{Functional Role:} This dimension acts as a structural consistency check. Even if color and components match, incorrect geometric relationships indicate a different product. The forbidden constraints specifically target unconventional designs (recumbent bikes, cargo bikes) that might share individual attributes but differ in overall configuration.

\subsubsection{Scene: Context Normalization}

\textbf{Required:}
\begin{itemize}[leftmargin=*,nosep]
    \item \textit{The product is displayed against a pure white or light-colored background.}
    \item \textit{The overall scene presents as an e-commerce or product catalog style clean photography environment.}
\end{itemize}

\textbf{Forbidden:}
\begin{itemize}[leftmargin=*,nosep]
    \item \textit{The product is displayed in a complex outdoor mountain environment.}
    \item \textit{The overall scene presents as a cluttered indoor space containing numerous other objects.}
\end{itemize}

\textbf{Ignore:}
\begin{itemize}[leftmargin=*,nosep]
    \item \textit{Minor brightness variations exist in the background.}
    \item \textit{Subtle shadow differences exist in ambient lighting.}
\end{itemize}

\textbf{Why These Constraints Emerge:} Scene evidence explicitly models the \textit{presentation context}, which is a major source of embedding noise. The same bicycle photographed in a studio versus on a mountain trail produces dramatically different embeddings, despite identical product identity.

\textbf{Functional Role:} By requiring studio-style presentation and ignoring minor background variations, this dimension normalizes context-induced similarity distortions. The forbidden constraints prevent matching to lifestyle photography where the product may be partially occluded or shown at unusual angles.

\subsubsection{Key Details: Discriminative Feature Anchoring}

\textbf{Required:}
\begin{itemize}[leftmargin=*,nosep]
    \item \textit{Brand logos are clearly printed on the frame side as an important identifying feature.}
    \item \textit{The bicycle is equipped with a color-matched chain guard and widened comfort saddle, reflecting daily riding orientation.}
\end{itemize}

\textbf{Forbidden:}
\begin{itemize}[leftmargin=*,nosep]
    \item \textit{Exaggerated racing numbers and competitive sponsor logos are printed on the frame.}
    \item \textit{The vehicle is equipped with dual-crown suspension forks and large-diameter disc brake systems as core features.}
\end{itemize}

\textbf{Ignore:}
\begin{itemize}[leftmargin=*,nosep]
    \item \textit{Minor differences exist in brand font sizing.}
    \item \textit{Slight positional variations exist in non-critical decals.}
\end{itemize}

\textbf{Why These Constraints Emerge:} Key details capture the \textit{most discriminative local features}—elements that uniquely identify a specific product or product line. Brand logos and signature components (color-matched chain guard) serve as ``visual fingerprints.''

\textbf{Functional Role:} This dimension handles the final disambiguation when multiple candidates pass all other checks. The forbidden constraints are particularly valuable: racing numbers and professional-grade suspension are strong negative indicators that immediately disqualify competitive cycling products, even if they share color and basic frame geometry.

\subsection{Integrated Analysis: How Evidence Dimensions Interact}

The six dimensions form a hierarchical filtering system:

\begin{enumerate}[leftmargin=*]
    \item \textbf{Entities} establishes category boundaries, eliminating cross-category confusion (mountain bikes, road bikes).
    \item \textbf{Attributes} performs intra-category discrimination based on visual properties (color, material, finish).
    \item \textbf{Actions} filters presentation state mismatches (assembled vs.\ disassembled).
    \item \textbf{Relations} verifies structural consistency with expected geometric configurations.
    \item \textbf{Scene} normalizes context variations that affect embedding similarity.
    \item \textbf{Key Details} anchors to unique identifying features for final disambiguation.
\end{enumerate}

The three-way constraint structure (required/forbidden/ignore) operates consistently across all dimensions:
\begin{itemize}[leftmargin=*]
    \item \textbf{Required}: Positive evidence that must be present—defines what the target \textit{is}.
    \item \textbf{Forbidden}: Negative evidence that must be absent—defines what the target \textit{is not}.
    \item \textbf{Ignore}: Variation tolerance—defines what \textit{does not matter} for identity.
\end{itemize}

This structured approach transforms retrieval from opaque similarity scoring into transparent logical evaluation, where each dimension's contribution can be independently assessed and explained.

\section{Prompt Templates}
\label{appendix:prompts}

This appendix presents the complete prompt templates used across the
\textbf{EviRank} pipeline. We employ four \emph{logical prompt modules}:
\textsc{P-Cap} for initial caption generation (Section~\ref{app:pcap}),
\textsc{P-B0} for query understanding and expansion (Section~\ref{app:pb0}),
\textsc{P-B1} for structured evidence slot mining (Section~\ref{app:pb1}),
and \textsc{P-Rank} for listwise re-ranking inference (Section~\ref{app:prank}).
The four modules are documented separately for clarity, but at inference
time they are merged into \emph{two online calls}:
\textsc{P-Cap}/\textsc{P-B0}/\textsc{P-B1} are completed within a single
structured evidence-extraction request, and \textsc{P-Rank} issues one
listwise call over the rubric top-$M$ candidates
(Section~\ref{sec:setup}).
All prompts are invoked in a training-free manner with no task-specific fine-tuning.
Placeholders enclosed in braces (e.g., \texttt{\{query\}}) are filled at runtime
with actual query-specific content.

\definecolor{pcapbg}{RGB}{234,246,255}
\definecolor{pb0bg}{RGB}{234,250,238}
\definecolor{pb1bg}{RGB}{255,244,229}
\definecolor{prankbg}{RGB}{245,234,255}

\subsection{Prompt P-Cap: Caption Generation}
\label{app:pcap}

\begin{figure*}[t]
\centering
\fcolorbox{black!25}{pcapbg}{%
\begin{minipage}{0.93\textwidth}
\small
\vspace{4pt}
\textbf{\textsc{Prompt P-Cap}~~~Detailed Caption Generation} \\[6pt]
You are an expert at generating comprehensive and detailed visual
descriptions for product images. Your task is to create a thorough
description that captures both the overall appearance and fine-grained
details of the product. \\[4pt]
\textbf{Query Image}: <attached> \\[4pt]
Generate a detailed product description that includes: \\[2pt]
\hspace{1em}1.~\textbf{Overall Appearance} -- Product type/category; general
shape and form factor; overall style (modern, vintage, classic, sporty, etc.). \\[1pt]
\hspace{1em}2.~\textbf{Color \& Material} -- Primary colors and any color
patterns/combinations; material composition (metal, plastic, leather, fabric,
wood, etc.); surface texture and finish (matte, glossy, brushed, etc.). \\[1pt]
\hspace{1em}3.~\textbf{Structural Details} -- Key components and parts;
construction and assembly details; size proportions and dimensions (relative
terms). \\[1pt]
\hspace{1em}4.~\textbf{Design Features} -- Unique design elements or patterns;
branding, logos, or text if present; decorative elements or embellishments.
\\[1pt]
\hspace{1em}5.~\textbf{Functional Elements} -- Visible functional parts;
hardware, fasteners, or connectors; any moving parts or adjustable features.
\\[4pt]
Please provide a comprehensive description in 3--5 detailed sentences that
would allow someone to accurately identify and distinguish this specific
product from similar items. Be as specific and descriptive as possible about
visual characteristics. \\[4pt]
\textbf{Output:} Only the detailed description text, nothing else.
\vspace{4pt}
\end{minipage}}
\caption{Prompt \textsc{P-Cap} for detailed caption generation.}
\label{fig:prompt_pcap}
\end{figure*}

Prompt \textsc{P-Cap} (Figure~\ref{fig:prompt_pcap}) serves as the entry
point of the evidence extraction pipeline. It converts a raw product image
into a rich, multi-aspect textual description covering five complementary
dimensions: overall appearance, color and material, structural details,
design features, and functional elements. This five-way decomposition is
intentionally designed to mirror the evidence-slot taxonomy introduced in
Section~\ref{sec:problem}, ensuring that the generated caption already
contains the fine-grained information needed for downstream evidence mining.
The prompt is particularly critical for image-to-image retrieval where no
user-provided text exists, as the caption becomes the sole textual anchor
for all subsequent stages. By requesting 3--5 detailed sentences, we ensure
sufficient specificity for discriminating among visually similar products
while keeping the output concise enough for efficient processing.

\subsection{Prompt P-B0: Query Understanding \& Expansion}
\label{app:pb0}

\begin{figure*}[t]
\centering
\fcolorbox{black!25}{pb0bg}{%
\begin{minipage}{0.93\textwidth}
\small
\vspace{4pt}
\textbf{\textsc{Prompt P-B0}~~~Query Understanding \& Expansion} \\[6pt]
You are a query understanding and expansion expert. Given a detailed image
caption/description, you need to:
(1)~Understand the core visual characteristics and product identity;
(2)~Provide contextual background and semantic expansion;
(3)~Output in a structured JSON format. \\[4pt]
\textbf{Input Caption}: {query}\\[4pt]
Please analyze and output in the following JSON format: \\[2pt]
\texttt{\{} \\
\hspace{1.5em}\texttt{"original\_query":~~"<original caption/query>",} \\
\hspace{1.5em}\texttt{"expanded\_query":~~"<expanded description with background context,} \\
\hspace{7.5em}\texttt{related concepts, and image-specific considerations>",} \\
\hspace{1.5em}\texttt{"core\_intent":~~~~~"<one sentence describing the main visual identity} \\
\hspace{7.5em}\texttt{and distinguishing features>",} \\
\hspace{1.5em}\texttt{"confidence":~~~~~~0.9} \\
\texttt{\}} \\[4pt]
\textbf{Requirements:} \\
$\bullet$~\texttt{expanded\_query} should add relevant visual and semantic
context while preserving all original details. \\
$\bullet$~\texttt{core\_intent} should capture the most distinctive visual
characteristics. \\
$\bullet$~\texttt{confidence} should be a number between 0.0 and 1.0. \\
$\bullet$~Rely only on the caption; do not use filenames, directory names,
or any other dataset-side metadata. \\
$\bullet$~Output ONLY the JSON object, no additional text before or after. \\
$\bullet$~Ensure all strings are properly escaped.
\vspace{4pt}
\end{minipage}}
\caption{Prompt \textsc{P-B0} for query understanding and expansion.}
\label{fig:prompt_pb0}
\end{figure*}

Prompt \textsc{P-B0} (Figure~\ref{fig:prompt_pb0}) enriches the raw caption
from the previous stage with background knowledge and semantic context
(Section~\ref{sec:mining}). The output contains three key fields.
\texttt{expanded\_query} augments the original description with concrete
visual details that are implicit in the caption---such as typical use
scenarios and related product concepts---thereby improving recall for
semantically relevant but lexically dissimilar candidates.
\texttt{core\_intent} distils the single most discriminative identity
statement, serving as the primary semantic anchor during evidence-slot
extraction. \texttt{confidence} is a self-assessed reliability estimate of
the expansion itself; it is used only to monitor and, if necessary, discard
degenerate expansions, and should not be confused with the listwise
confidence $p$ that weights the distillation loss in Stage~C
(Eq.~\ref{eq:total_loss}), which is produced by \textsc{P-Rank}. The prompt
is conditioned on the caption alone: image paths, filenames and directory
names are never exposed to the teacher at any stage, so no category prior
can leak from the dataset layout into the evidence package.

\subsection{Prompt P-B1: Structured Evidence Slot Mining}
\label{app:pb1}

\begin{figure*}[t]
\centering
\fcolorbox{black!25}{pb1bg}{%
\begin{minipage}{0.93\textwidth}
\small
\vspace{4pt}
\textbf{\textsc{Prompt P-B1}~~~Structured Evidence Slot Mining} \\[6pt]
You are a visual evidence extraction expert for image-to-image retrieval.
Given a detailed image caption, extract structured evidence that would
help identify matching images. \\[4pt]
\textbf{Input Caption}: {query} \\[4pt]
Your task is to extract evidence into \textbf{6 semantic slots}, with each
slot containing 2--4 complete \textbf{affirmative sentences} (not phrases).
Organize into Required\,/\,Forbidden\,/\,Ignore lists: \\[2pt]
$\bullet$~\textbf{Required}: Affirmative sentences describing visual elements
that MUST be present for a correct match. \\
\hspace{1.5em}\textit{Example: ``The bicycle is blue with silver handlebars.''} \\[1pt]
$\bullet$~\textbf{Forbidden}: Affirmative sentences describing visual elements
that would indicate a WRONG match. These describe what incorrect matches
would look like. \\
\hspace{1.5em}\textit{Example: ``The bicycle is red with black handlebars.''}
\\[1pt]
$\bullet$~\textbf{Ignore}: Affirmative sentences describing similar but
non-critical visual elements that can be tolerated. \\[4pt]
\textit{Scoring logic:}~~$\text{score}=\sum\text{(required\_match\_scores)}
-\sum\text{(forbidden\_match\_scores)}$ \\[4pt]
Output in the following JSON format: \\[2pt]
\texttt{\{} \\
\hspace{1em}\texttt{"global\_summary": "<2--3 sentences providing an overall visual summary>",} \\
\hspace{1em}\texttt{"entities":~~~~\{"required":[\ldots], "forbidden":[\ldots], "ignore":[\ldots]\},} \\
\hspace{1em}\texttt{"attributes":~~\{"required":[\ldots], "forbidden":[\ldots], "ignore":[\ldots]\},} \\
\hspace{1em}\texttt{"actions":~~~~~\{"required":[\ldots], "forbidden":[\ldots], "ignore":[\ldots]\},} \\
\hspace{1em}\texttt{"relations":~~~\{"required":[\ldots], "forbidden":[\ldots], "ignore":[\ldots]\},} \\
\hspace{1em}\texttt{"scene":~~~~~~~\{"required":[\ldots], "forbidden":[\ldots], "ignore":[\ldots]\},} \\
\hspace{1em}\texttt{"key\_details":~\{"required":[\ldots], "forbidden":[\ldots], "ignore":[\ldots]\}} \\
\texttt{\}} \\[4pt]
\textbf{Requirements:} \\
$\bullet$~\textbf{All} sentences must be affirmative (positive statements),
including forbidden items. \\
$\bullet$~Forbidden sentences describe what wrong matches look like (positive
form), NOT using ``not'' or ``should not''. \\
$\bullet$~Each entry should be a complete sentence; each list should have
2--4 sentences. \\
$\bullet$~Base the evidence strictly on the visual content described in the caption; do not rely on filenames, directories, or any external metadata. \\
$\bullet$~Sentences should be specific and descriptive for accurate embedding
matching. \\
$\bullet$~Output ONLY valid JSON, no markdown code blocks, no explanations.
\vspace{4pt}
\end{minipage}}
\caption{Prompt \textsc{P-B1} for structured evidence slot mining.}
\label{fig:prompt_pb1}
\end{figure*}

Prompt \textsc{P-B1} (Figure~\ref{fig:prompt_pb1}) is the core
evidence-mining prompt. It
decomposes the enriched query into six orthogonal semantic
dimensions---\textsc{Entities}, \textsc{Attributes}, \textsc{Actions},
\textsc{Relations}, \textsc{Scene}, and \textsc{KeyDetails}---each further
organized into \emph{required}, \emph{forbidden}, and \emph{ignore}
constraint lists. Two design choices are noteworthy.
\textbf{(i)~Affirmative-only wording.} All constraint sentences, including
those in the forbidden list, are phrased as positive statements that
\emph{describe what a wrong match looks like} rather than negating the
target. For instance, instead of ``The bicycle should not be red,'' the
forbidden constraint states ``The bicycle is red with black handlebars.''
This ensures that every constraint can be directly compared with candidate
descriptions via embedding cosine similarity
(Eqs.~\ref{eq:slot_rates}--\ref{eq:rubric}), yielding stable
and sign-consistent matching scores.
\textbf{(ii)~Explicit scoring guidance.} The prompt communicates the
intended scoring logic
($\text{score}=\sum\text{required\_match}-\sum\text{forbidden\_match}$),
aligning constraint generation with the downstream deterministic rubric
(Eq.~\ref{eq:rubric}) and reducing misalignment between extraction and evaluation.

\subsection{Prompt P-Rank: Listwise Re-ranking}
\label{app:prank}

During inference, the teacher MLLM receives both the query image and all
top-$K$ candidate images simultaneously, together with the structured
evidence text distilled from the preceding stages. The evidence text
presented to the re-ranker is assembled at runtime from the outputs of
\textsc{P-B0} and \textsc{P-B1} following a compact serialization format
(described after the prompt box). Figure~\ref{fig:prompt_prank} presents
the full re-ranking prompt.

\begin{figure*}[htpb]
\centering
\fcolorbox{black!25}{prankbg}{%
\begin{minipage}{0.93\textwidth}
\small
\vspace{4pt}
\textbf{\textsc{Prompt P-Rank}~~~Listwise Re-ranking with Evidence Constraints and Hard-Pair Mining} \\[6pt]

You are a visual reranker for image-to-image product retrieval.
Given a query image, its detailed description, structured evidence constraints,
and a list of candidate images, rank the candidate images from best to worst
match based on visual similarity to the query image. \\[4pt]

The goal is to find images of the SAME product or very similar products as
the query image. Focus on: \\[1pt]
\hspace{1em}$\bullet$~Exact or near-exact visual matches, including the same product from different angles. \\
\hspace{1em}$\bullet$~Products with identical key visual features, such as shape, color, branding, and distinctive details. \\
\hspace{1em}$\bullet$~Products from the same product line or category. \\[4pt]

In addition to the final ranking, report (i)~an \textbf{absolute relevance
score} for every candidate, (ii)~a \textbf{self-assessed confidence} in the
ordering, and (iii)~\textbf{hard-pair cases}: candidate pairs whose relevance
scores are very close or visually difficult to distinguish. These scores,
the confidence and the hard pairs are used as fine-grained distillation
signals for training the student re-ranker. \\[4pt]

\textbf{IMPORTANT OUTPUT RULES:} \\
$\bullet$~Output MUST be a single JSON object. \\
$\bullet$~JSON schema: \\[1pt]
\hspace{1.5em}\texttt{\{} \\
\hspace{2.5em}\texttt{"ranked\_indices": [int, int, \ldots],} \\
\hspace{2.5em}\texttt{"relevance\_scores": [} \\
\hspace{3.5em}\texttt{\{"index": int, "score": float\}} \\
\hspace{2.5em}\texttt{],} \\
\hspace{2.5em}\texttt{"confidence": float,} \\
\hspace{2.5em}\texttt{"hard\_pairs": [} \\
\hspace{3.5em}\texttt{\{"pair": [int, int], "reason": "short reason"\}} \\
\hspace{2.5em}\texttt{],} \\
\hspace{2.5em}\texttt{"brief\_reasoning": "one short paragraph explaining the ranking"} \\
\hspace{1.5em}\texttt{\}} \\[1pt]
$\bullet$~\texttt{"ranked\_indices"} must be a permutation of
$0\,..\,\{N{-}1\}$, ordered from BEST match to WORST match. \\
$\bullet$~\texttt{"relevance\_scores"} must contain one entry per candidate
index, with \texttt{score} on a 0--100 scale and consistent with
\texttt{ranked\_indices}. \\
$\bullet$~\texttt{"confidence"} must be a single number between 0.0 and 1.0
expressing how decisively the candidates can be ordered under the given
evidence. \\
$\bullet$~\texttt{"hard\_pairs"} should contain visually ambiguous candidate pairs
with close relevance to the query. \\
$\bullet$~Do NOT output any extra text outside JSON. \\[6pt]

\texttt{=== QUERY IMAGE DESCRIPTION ===} \\
\texttt{\{generated\_caption\}} \\[4pt]

\texttt{=== EVIDENCE CONSTRAINTS ===} \\
\texttt{\{evidence\_text\}} \\[4pt]

\texttt{=== CANDIDATE LIST (index -> opaque ID) ===} \\
\texttt{\{candidates\_desc\}}\\[4pt]

\textnormal{Below~you~will~see:} \\[2pt]
1.~The~\textbf{QUERY IMAGE}~(labeled)~\quad---\quad
this~is~what~you~are~trying~to~match \\
2.~The~\textbf{CANDIDATE IMAGES}~in~order~$(0,1,2,\ldots)$~\quad---\quad
rank~these~by~similarity~to~the~query \\[4pt]

\texttt{[QUERY IMAGE]}~~$\langle$\textit{base64-encoded query image}$\rangle$ \\[2pt]
\texttt{[CANDIDATE IMAGES]}~~$\langle$\textit{base64-encoded candidate images, indexed 0..N\!-\!1}$\rangle$

\vspace{4pt}
\end{minipage}}
\caption{Prompt \textsc{P-Rank} for listwise re-ranking inference and hard-pair mining.
The evidence text block is assembled from \textsc{P-B0} and \textsc{P-B1}
outputs as described in Section~\ref{app:evidence_assembly}. Hard-pair cases
are candidate pairs with close relevance scores or subtle visual differences,
which are further used as fine-grained distillation signals for training the
student re-ranker.}
\label{fig:prompt_prank}
\end{figure*}
The re-ranking prompt is designed around three principles.
\textbf{(i)~Evidence-conditioned ranking.} The evidence constraints
(\texttt{\{evidence\_text\}}) are placed prominently before the candidate
list, ensuring that the teacher attends to structured required/forbidden
constraints as the primary decision basis rather than relying solely on
holistic visual impression. This directly implements the
``rubric-first stabilization'' strategy.
\textbf{(ii)~Multimodal listwise input.} Both the query image and all $K$
candidate images are presented in a single prompt turn, enabling the teacher
to perform direct visual comparison across candidates rather than scoring
each independently. The images are base64-encoded with controlled
resolution (\texttt{max\_edge}$=$2048) and JPEG quality (\texttt{quality}$=$85)
to balance visual fidelity and token efficiency.
\textbf{(iii)~Structured JSON output.} By requiring a strict JSON schema
with a permutation of candidate indices, we eliminate ambiguity in parsing
and guarantee a valid ranking. The same call also returns the three
supervision fields consumed by Stage~C: \texttt{relevance\_scores} provide
the teacher scores $\mathbf{r}$ used in $\mathcal{L}_{\text{score}}$,
\texttt{confidence} provides the weight $p$ in Eq.~\ref{eq:total_loss}, and
\texttt{hard\_pairs} provides $\mathcal{H}$ for
$\mathcal{L}_{\text{pair}}$; only \texttt{ranked\_indices} is needed to
produce the final ranking at test time. The accompanying
\texttt{brief\_reasoning} field, while not used in the final score, provides
interpretable rationales that are valuable for error analysis.

We anonymize candidate identifiers (replacing filenames with
opaque indices) before presenting them to the teacher, so that
no dataset-side metadata reaches the re-ranker. For large candidate sets ($K > 8$), we adopt a \emph{divide-and-merge}
batching strategy: candidates are partitioned into groups of size $B$
(default $B{=}5$), each group is independently ranked by the teacher, and
the resulting sub-lists are iteratively merged via pairwise re-ranking
calls. With $n=\lceil K/B\rceil$ groups this costs $2n-1$ calls in
total and $1+\lceil\log_2 n\rceil$ sequential rounds when merges at
the same tree level are issued in parallel, which keeps inference
cost tractable while preserving global ranking quality.

\subsection{Evidence Text Assembly}
\label{app:evidence_assembly}

The \texttt{\{evidence\_text\}} placeholder in \textsc{P-Rank} is populated
at runtime by serializing the outputs of \textsc{P-B0} and \textsc{P-B1}
into a compact textual format. Figure~\ref{fig:evidence_assembly} shows
the assembly template.

\definecolor{assemblybg}{RGB}{245,245,250}

\begin{figure*}[t]
\centering
\fcolorbox{black!25}{assemblybg}{%
\begin{minipage}{0.93\textwidth}
\small
\vspace{4pt}
\textbf{Evidence Text Assembly Template} \\[6pt]
\texttt{Core intent: \{b0.core\_intent\}} \\[2pt]
\texttt{Expanded query: \{b0.expanded\_query\}} \\[2pt]
\texttt{Global summary: \{b1.global\_summary\}} \\[2pt]
\texttt{Slot constraints:} \\
\texttt{~~[entities]} \\
\texttt{~~~~required: \{sent\_1\} | \{sent\_2\} | \ldots} \\
\texttt{~~~~forbidden: \{sent\_1\} | \{sent\_2\} | \ldots} \\
\texttt{~~[attributes]} \\
\texttt{~~~~required: \{sent\_1\} | \{sent\_2\} | \ldots} \\
\texttt{~~~~forbidden: \{sent\_1\} | \{sent\_2\} | \ldots} \\
\texttt{~~[actions]} \\
\texttt{~~~~required: \ldots ~~~~forbidden: \ldots} \\
\texttt{~~[relations]} \\
\texttt{~~~~required: \ldots ~~~~forbidden: \ldots} \\
\texttt{~~[scene]} \\
\texttt{~~~~required: \ldots ~~~~forbidden: \ldots} \\
\texttt{~~[key\_details]} \\
\texttt{~~~~required: \ldots ~~~~forbidden: \ldots}
\vspace{4pt}
\end{minipage}}
\caption{The runtime assembly format for evidence text injected into
\textsc{P-Rank}. Constraint sentences within each slot are concatenated
with pipe delimiters (\texttt{|}). Only \emph{required} and
\emph{forbidden} constraints are included; \emph{ignore} constraints are
omitted from the re-ranking prompt as they serve primarily to calibrate
the deterministic rubric score (Eq.~\ref{eq:rubric}) rather than guide the MLLM.}
\label{fig:evidence_assembly}
\end{figure*}
As shown in Figure~\ref{fig:evidence_assembly}, the assembled evidence text
follows a three-tier hierarchy: \emph{query-level} summaries (core intent,
expanded query, global summary), followed by \emph{slot-level} constraint
blocks. Within each slot, required and forbidden constraint sentences are
serialized as pipe-delimited strings, with a maximum of six sentences per
list to control prompt length. Notably, the \emph{ignore} constraints are
deliberately excluded from the re-ranking prompt. This is because their
role is to calibrate the deterministic rubric scoring (Eq.~\ref{eq:rubric}) by masking
non-discriminative variations, whereas the MLLM re-ranker is expected to
internalize such tolerance through its own visual reasoning. Including
ignore constraints would increase prompt length without providing
actionable ranking guidance. This selective serialization keeps the
evidence text compact (typically 200--400 tokens) while preserving all
decision-critical information.

\medskip

Together, the four prompts form a progressive refinement pipeline:
\textsc{P-Cap} grounds the image in language, \textsc{P-B0} enriches the
description with semantic context, \textsc{P-B1} structures the enriched
representation into checkable constraints, and \textsc{P-Rank} leverages
the full evidence package for listwise visual re-ranking. Each stage builds
upon the previous one's output, creating a coherent chain from raw pixels
to interpretable, evidence-conditioned ranking decisions.

\section{Slot Decomposition Example} \label{slot_decomposition}

The brief slot template is given as Table~\ref{tab:slot_decomposition}.

\begin{table*}[t]
\centering
\small
\setlength{\tabcolsep}{6pt}
\renewcommand{\arraystretch}{1.25}
\begin{tabularx}{\textwidth}{@{}>{\bfseries}p{1.7cm} X X X @{}}
\toprule
Slot & Required & Forbidden & Ignore \\
\midrule
Entities &
The image must contain an acoustic guitar with a hollow wooden body and a visible sound hole. The guitar should have a traditional construction with a neck and fretboard clearly visible. &
The guitar must not be an electric guitar with solid body and pickups. The instrument should not be a ukulele, bass guitar, banjo, or any other stringed instrument. &
The specific brand or manufacturer of the guitar is not relevant for matching. The exact number of strings or whether a pick is being used does not affect the match. \\
\addlinespace

Attributes &
The dress must have blue as its primary base color. The fabric must display a floral or flower-based pattern with visible botanical motifs. &
The dress should not be a solid single color without any pattern. The pattern must not be geometric stripes, polka dots, or abstract non-floral designs. &
The exact shade or tone of blue does not need to match precisely. The specific type of flowers in the pattern and the overall dress length are not critical factors. \\
\addlinespace

Actions &
The woman must be actively engaged in playing the guitar with fingers positioned on strings. The posture should indicate ongoing musical performance rather than static positioning. &
The woman should not be merely holding the guitar as a prop without playing engagement. The scene must not show her posing statically for a photograph with the guitar as an accessory. &
The specific playing technique such as strumming versus fingerpicking is not relevant. The particular song or music being performed does not affect the matching. \\
\addlinespace

Relations &
The woman's hands must be positioned on the guitar in a playing configuration, with one hand on the neck and one near the body. The guitar must be held in an active playing position against her body. &
The guitar should not be lying flat on a surface or hanging on a wall. The guitar must not be stored in a case or held passively without playing intent. &
Whether the woman is sitting or standing while playing is not relevant. The specific hand positioning for left-handed versus right-handed playing does not affect the match. \\
\addlinespace

Scene &
The setting must be an outdoor café environment with visible tables, chairs, or typical café furniture. The lighting should indicate an outdoor daytime setting. &
The location must not be an indoor concert stage or formal performance venue. The setting should not be a recording studio, private bedroom, or enclosed indoor space. &
The number of other patrons or customers visible in the scene is not relevant. The specific café name, branding, or current weather conditions do not affect the match. \\
\addlinespace

KeyDetails &
The overall atmosphere must suggest a casual, informal musical performance in a relaxed social setting. The context should indicate spontaneous or low-key entertainment rather than organized performance. &
The scene must not depict a formal concert with audience seating arrangements. The context should not suggest a music competition, audition, or professional street busking for monetary collection. &
The specific time of day or season when the scene takes place is not relevant. The geographic location or cultural context of the café does not affect the matching. \\
\bottomrule
\end{tabularx}
\caption{Slot Decomposition: structured evidence slots with required, forbidden, and ignored constraints.}
\label{tab:slot_decomposition}
\end{table*}

\section{Robustness Analysis Details}
\label{appendix:robustness}

We complement the main-text robustness and efficiency summaries
(Sections~\ref{sec:robustness} and~\ref{sec:efficiency}) with full tables on
FashionIQ (mean over Shirt/Dress/Toptee).

\paragraph{(a) Repeated-run stability.}
We invoke the same teacher API 10$\times$ with identical inputs and perturb the
student's random seed / decoding temperature.
\begin{center}
\small
\begin{adjustbox}{max width=\linewidth}
\begin{tabular}{lccc}
\toprule
\textbf{Setting} & \textbf{R@10} & \textbf{Kendall's $\tau$} & \textbf{Top-1 Agr.} \\
\midrule
Repeated calls       & 42.93$\pm$0.27 & 0.99 & 99.6\% \\
Seed perturbation    & 32.03$\pm$1.26 & 0.93 & 98.2\% \\
Decoding randomness  & 32.13$\pm$0.89 & 0.96 & 98.8\% \\
\bottomrule
\end{tabular}
\end{adjustbox}
\end{center}

\paragraph{(b) Prompt perturbation.}
We perturb the evidence-extraction prompt with three transformations.
\begin{center}\small
\begin{adjustbox}{max width=\linewidth}
\begin{tabular}{@{}lccc@{}}
\toprule
\textbf{Variant} & \textbf{R@10} & \textbf{Kendall's $\tau$} & \textbf{Top-1 Agr.} \\
\midrule
Paraphrase      & 42.6 & 0.97 & 98.8\% \\
Order-Shuffled  & 42.4 & 0.95 & 99.3\% \\
Minor Typos     & 41.9 & 0.95 & 98.8\% \\
\bottomrule
\end{tabular}
\end{adjustbox}
\end{center}

\paragraph{(c) Cross-teacher.}
We replace the teacher MLLM with five different models.
\begin{center}\small
\begin{adjustbox}{max width=\linewidth}
\begin{tabular}{@{}lccc@{}}
\toprule
\textbf{Teacher} & \textbf{R@10} & \textbf{Kendall's $\tau$} & \textbf{Top-1 Agr.} \\
\midrule
Gemini-3-pro       & 42.9 & 1.00 & 100\% \\
GPT-5.4            & 41.3 & 0.92 & 92.0\% \\
Claude-4.6-Opus    & 42.7 & 0.91 & 94.2\% \\
Claude-4.6-Sonnet  & 40.9 & 0.89 & 93.3\% \\
Gemini-3-flash     & 38.3 & 0.90 & 91.1\% \\
\bottomrule
\end{tabular}
\end{adjustbox}
\end{center}

R/F/I label agreement across runs is $\geq$99\% in all settings, confirming the
structured-slot output format is highly stable.

\section{Prompt Sensitivity and Information-Source Ablation}
\label{appendix:prompt_sens}

These experiments verify that EviRank's gains stem from structured-evidence
slots rather than prompt engineering.

\paragraph{Free-form augmentation vs.\ structured evidence.}
\begin{center}
\small
\setlength{\tabcolsep}{3pt}
\begin{adjustbox}{max width=\linewidth}
\begin{tabular}{lcccc}
\toprule
\textbf{Setting} & \textbf{COCO R@1} & \textbf{SoP R@1} & \textbf{FIQ R@10} & \textbf{FIQ R@50} \\
\midrule
Query-only listwise         & 63.02 & 84.73 & 36.57 & 54.43 \\
Free-form augmentation      & 63.53 & 86.70 & 38.29 & 56.72 \\
Structured evidence (ours)  & 69.53 & 91.46 & 42.90 & 60.41 \\
\bottomrule
\end{tabular}
\end{adjustbox}
\end{center}

\paragraph{Information-source ablation.}
\begin{center}
\small
\setlength{\tabcolsep}{3pt}
\begin{adjustbox}{max width=\linewidth}
\begin{tabular}{lccc}
\toprule
\textbf{Setting} & \textbf{COCO R@1} & \textbf{SoP R@1} & \textbf{FIQ R@10} \\
\midrule
Original query only          & 63.02 & 84.73 & 36.57 \\
Query + caption only         & 64.79 & 85.28 & 37.93 \\
Query + evidence only        & 68.17 & 90.83 & 42.19 \\
Query + caption + evidence   & 69.53 & 91.46 & 42.90 \\
\bottomrule
\end{tabular}
\end{adjustbox}
\end{center}

\paragraph{Prompt template sensitivity.}
\begin{center}
\small
\setlength{\tabcolsep}{3pt}
\begin{adjustbox}{max width=\linewidth}
\begin{tabular}{lccc}
\toprule
\textbf{Template} & \textbf{COCO R@1} & \textbf{SoP R@1} & \textbf{FIQ R@10} \\
\midrule
A: Original prompt              & 69.53 & 91.46 & 42.90 \\
B: Concise version              & 69.12 & 91.24 & 42.65 \\
C: Rewritten version            & 69.50 & 91.38 & 42.83 \\
D: No explicit penalty wording  & 68.86 & 90.72 & 42.12 \\
\bottomrule
\end{tabular}
\end{adjustbox}
\end{center}
The variation across rewrites is well below the gap from removing structured
evidence, confirming that improvements come from the slot-wise R/F/I formulation
rather than prompt wording.

\section{Slot Semantics Analysis}
\label{appendix:slot_analysis}

This appendix consolidates the semantic-independence analysis and the slot-type
functional analysis previously placed in the main text of the original
manuscript.

\subsection{Semantic Independence Across Slots}

\begin{figure}[h]
  \centering
  \includegraphics[width=\columnwidth]{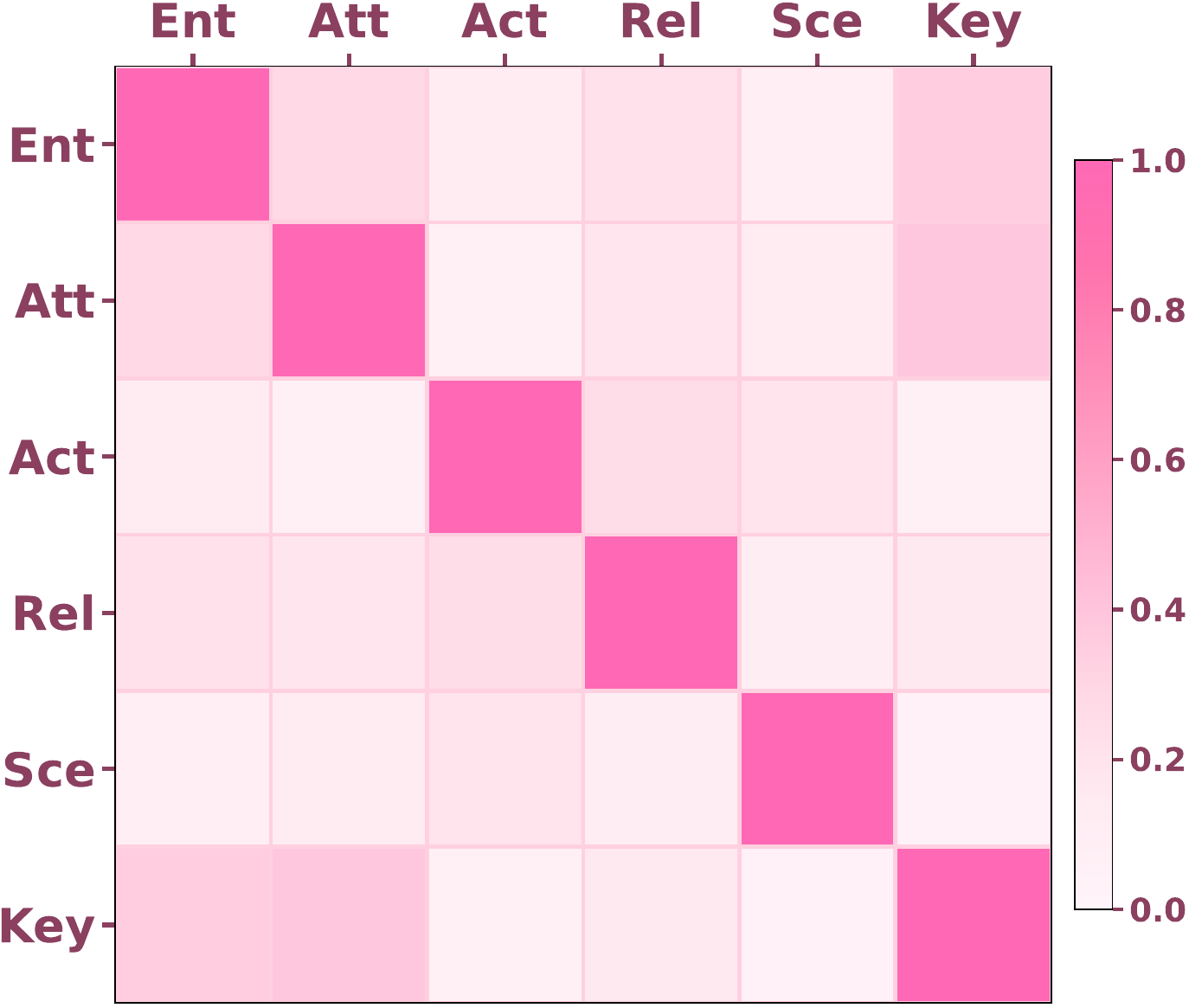}
  \caption{Semantic similarity matrix across evidence slots.}
  \label{fig:heatmap}
\end{figure}

We compute pairwise semantic similarity using Sentence-BERT embeddings across
extracted constraints from 10k randomly sampled queries. Figure~\ref{fig:heatmap}
presents the similarity matrix; low off-diagonal values (mean = 0.18) confirm
semantic orthogonality. The lowest similarities occur between
\textit{Scene}--\textit{KeyDetails} (0.05) and \textit{Actions}--\textit{Attributes}
(0.06), reflecting fundamentally different semantic axes: environmental context
vs.\ product identity, and dynamic state vs.\ static appearance. Moderate
correlations between \textit{Entities}--\textit{KeyDetails} (0.35) and
\textit{Attributes}--\textit{KeyDetails} (0.38) reveal a semantic hierarchy
rather than redundancy: \textit{Entities} establishes category-level identity
while \textit{KeyDetails} anchors instance-level discrimination; \textit{Attributes}
captures global appearance while \textit{KeyDetails} identifies unique local
features. This validates our principle that slots should be \emph{semantically
orthogonal yet functionally complementary}.

\subsection{Functional Roles of Slot Types}

\begin{figure}[h]
  \centering
  \includegraphics[width=0.98\columnwidth]{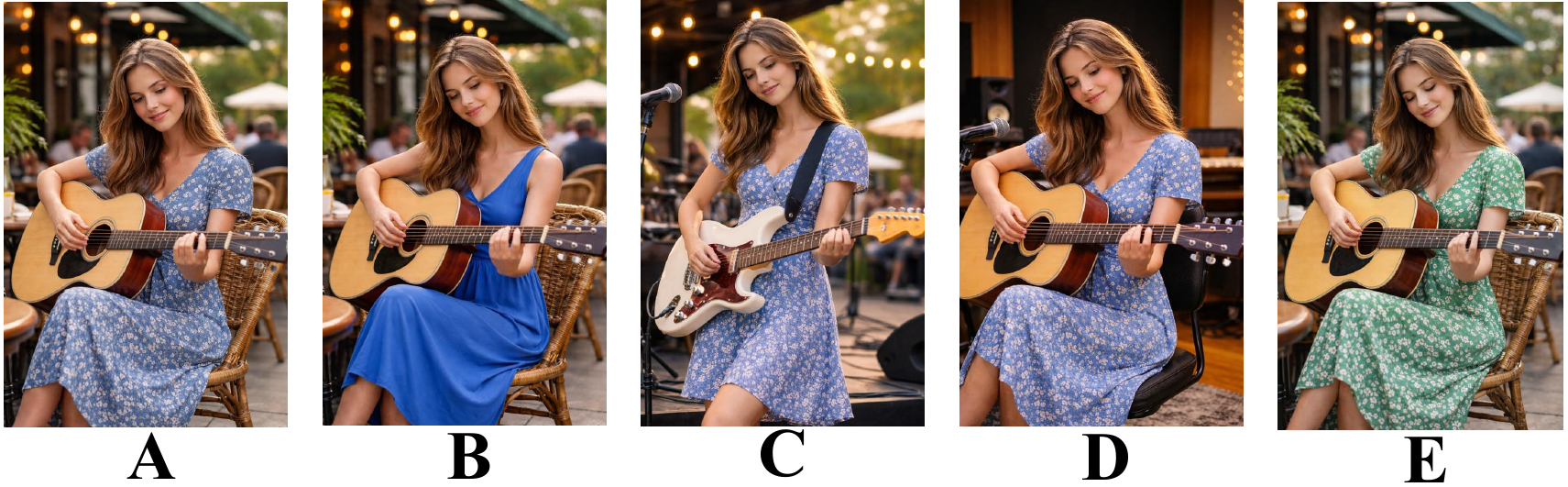}
  \caption{Candidate images for slot-type ablation.}
  \label{fig:slot}
\end{figure}

\begin{table}[h]
\centering
\small
\caption{Ranking shifts under evidence-constraint ablation.}
\label{tab:slot}
\setlength{\tabcolsep}{3pt}
\renewcommand{\arraystretch}{1.1}
\begin{adjustbox}{max width=\columnwidth}
\begin{tabular}{@{}lcccc@{}}
\toprule
\textbf{Candidate} & \textbf{Full} & \textbf{w/o R} & \textbf{w/o F} & \textbf{w/o I} \\
\midrule
(A) Blue floral dress, acoustic, caf\'e   & 1 & 4 & 1 & 1 \\
(B) Blue solid dress, acoustic, caf\'e    & 3 & 2 & 1 & 3 \\
(C) Blue floral dress, electric, stage    & 5 & 3 & 3 & 5 \\
(D) Blue floral dress, acoustic, studio   & 4 & 5 & 4 & 2 \\
(E) Green floral dress, acoustic, caf\'e  & 2 & 1 & 2 & 4 \\
\bottomrule
\end{tabular}
\end{adjustbox}
\end{table}

To illustrate how different constraint types influence the re-ranking process,
we analyze the query: \emph{``A young woman in a blue floral dress playing
acoustic guitar at an outdoor caf\'e.''} The structured evidence slots for this
query decompose as given in Table~\ref{tab:slot_decomposition} (Appendix~\ref{slot_decomposition}).
Figure~\ref{fig:slot} and Table~\ref{tab:slot} show how candidate rankings shift
when specific constraint types are ablated. Without required constraints,
candidate E (green dress) rises to rank 1 since blue is no longer anchored.
Without forbidden constraints, candidates B (solid pattern) and C (electric
guitar) gain unjustified ground. Without ignore constraints, candidate D
(indoor studio) is over-promoted since setting variations are no longer
filtered, while candidate E drops to rank 4 because the green-vs-blue
difference is now weighted more heavily without shade variations being
explicitly ignorable. This three-way decomposition transforms re-ranking from
holistic similarity scoring into explicit multi-dimensional logical evaluation,
which additionally provides structured and decomposable supervision signals
that facilitate knowledge distillation to smaller models.

\section{Correspondence to SQL Constraints}
\label{appendix:sql}

Reviewer feedback noted an interesting structural correspondence between our
required/forbidden/ignore (R/F/I) labels and classical SQL constraint
mechanisms. While SQL constraints are \emph{schema-level} rules attached to a
fixed relational model, ours are \emph{query-time} semantic rubrics generated
per query. Nonetheless, the mapping provides intuitive grounding:

\begin{center}\small
\setlength{\tabcolsep}{4pt}
\begin{adjustbox}{max width=\linewidth}
\begin{tabular}{@{}lll@{}}
\toprule
\textbf{EviRank} & \textbf{SQL analog} & \textbf{Role} \\
\midrule
Required  & \texttt{NOT NULL} / mandatory predicate & must hold \\
Forbidden & negated \texttt{CHECK} constraint       & must not hold \\
Ignore    & \texttt{NULL}-able / unconstrained      & no constraint \\
\bottomrule
\end{tabular}
\end{adjustbox}
\end{center}

This mapping highlights that EviRank effectively performs constraint
satisfaction over a query-conditioned rubric, analogous to schema-validated
record matching but with the ``schema'' dynamically generated by the MLLM
teacher.

\section{Detailed Comparison with ImageScope}
\label{appendix:imagescope_compare}

ImageScope~\cite{Luo2025ImageScope} is the most closely related
multi-stage MLLM-based re-ranker, and it likewise unifies
heterogeneous queries before invoking an MLLM. We therefore make
the formulation-level differences explicit.

\paragraph{Different unification targets.}
ImageScope unifies queries into a \emph{single natural-language
description} of the user's intent and lets the MLLM perform
holistic reasoning over that description. The unified
representation is therefore still a piece of free-form text, and
the relevance decision is recovered \emph{implicitly} from the
MLLM's reasoning trace. EviRank, in contrast, unifies queries
into a \emph{typed evidence package} $\mathcal{E}(q)$ consisting
of slot-wise required/forbidden/ignorable constraints. The
unified representation is itself an evaluation rubric, and
relevance is recovered \emph{explicitly} via constraint
verification rather than holistic reasoning.

\paragraph{Different failure surfaces.}
Because ImageScope's representation is a holistic description
followed by free-form reasoning, it inherits the three failure
modes of CoT-based re-rankers discussed in
Section~\ref{sec:related} (omission of unstated constraints,
hallucinated evidence, and inconsistent semantic coverage across
queries). EviRank's typed-evidence representation directly
addresses these failure modes: required and forbidden statements
force the model to commit to which dimensions matter
\emph{before} comparing candidates; ignorable statements
explicitly model what should not be penalized; and the slot
schema is fixed across queries, making coverage uniform and
auditable.

\paragraph{Auditability and supervision.}
Holistic descriptions are not decomposable: it is hard to ask
``which part of the description is satisfied by candidate $c$?''
or to extract per-dimension supervision signals from a single
free-form passage. EviRank's evidence package is decomposable by
construction, which (i) yields per-slot satisfaction and
violation rates that can be inspected and verified independently,
and (ii) provides structured supervision (slot-wise indicators,
hard pairs, calibrated scores) that can be distilled into a
smaller student---neither of which is naturally available from
holistic-description pipelines.

\paragraph{Empirical gap.}
The two formulations are not just stylistically different. Across
all five benchmarks (Tables~\ref{tab:flickr30k}--\ref{tab:fashioniq}),
EviRank-pro consistently outperforms ImageScope by large margins,
and even the rubric-only EviRank-mini---which uses no MLLM at
test time---matches or exceeds ImageScope, indicating that the
gain originates from the structured-evidence representation
itself rather than from additional MLLM compute.

\end{document}